\documentclass[conference]{IEEEtran}
\IEEEoverridecommandlockouts
\usepackage{cite}
\usepackage{amsmath,amssymb,amsfonts,bm}
\usepackage{algorithmic}
\usepackage{float}
\usepackage{graphicx}
\usepackage{textcomp}
\usepackage{xcolor}
\usepackage{subfigure}
\usepackage[ruled]{algorithm2e}
\def\BibTeX{{\rm B\kern-.05em{\sc i\kern-.025em b}\kern-.08em
    T\kern-.1667em\lower.7ex\hbox{E}\kern-.125emX}}

\usepackage[colorlinks, linkcolor=blue, anchorcolor=blue, citecolor=blue]{hyperref}
\begin{document}

\title{Deep Monte Carlo Quantile Regression for Quantifying Aleatoric Uncertainty in Physics-informed Temperature Field Reconstruction}

\author{\IEEEauthorblockN{1\textsuperscript{st} Xiaohu Zheng}
\IEEEauthorblockA{\textit{College of Aerospace Science and Engineering} \\
\textit{National University of Defense Technology}\\
Changsha, China \\
zhengboy320@163.com}
\and
\IEEEauthorblockN{2\textsuperscript{rd} Wen Yao}
\IEEEauthorblockA{\textit{Defense Innovation Institute} \\
\textit{Chinese Academy of Military Science}\\
Beijing, China \\
wendy0782@126.com}
\and
\IEEEauthorblockN{3\textsuperscript{nd} Zhiqiang Gong}
\IEEEauthorblockA{\textit{Defense Innovation Institute} \\
\textit{Chinese Academy of Military Science}\\
Beijing, China \\
gongzhiqiang13@nudt.edu.cn}
\and
\IEEEauthorblockN{4\textsuperscript{th} Yunyang Zhang}
\IEEEauthorblockA{\textit{Defense Innovation Institute} \\
\textit{Chinese Academy of Military Science}\\
Beijing, China \\
zhangyunyang17@csu.ac.cn}
\and
\IEEEauthorblockN{5\textsuperscript{th} Xiaoyu Zhao}
\IEEEauthorblockA{\textit{Defense Innovation Institute} \\
\textit{Chinese Academy of Military Science}\\
Beijing, China \\
172826256@qq.com}
\and
\IEEEauthorblockN{6\textsuperscript{th} Tingsong Jiang}
\IEEEauthorblockA{\textit{Defense Innovation Institute} \\
\textit{Chinese Academy of Military Science}\\
Beijing, China \\
tingsongad@163.com}
}

\maketitle

\begin{abstract}
For the temperature field reconstruction (TFR), a complex image-to-image regression problem, the convolutional neural network (CNN) is a powerful surrogate model due to the convolutional layer's good image feature extraction ability. However, a lot of labeled data is needed to train CNN, and the common CNN can not quantify the aleatoric uncertainty caused by data noise. In actual engineering, the noiseless and labeled training data is hardly obtained for the TFR. To solve these two problems, this paper proposes a deep Monte Carlo quantile regression (Deep MC-QR) method for reconstructing the temperature field and quantifying aleatoric uncertainty caused by data noise. On the one hand, the Deep MC-QR method uses physical knowledge to guide the training of CNN. Thereby, the Deep MC-QR method can reconstruct an accurate TFR surrogate model without any labeled training data. On the other hand, the Deep MC-QR method constructs a quantile level image for each input in each training epoch. Then, the trained CNN model can quantify aleatoric uncertainty by quantile level image sampling during the prediction stage. Finally, the effectiveness of the proposed Deep MC-QR method is validated by many experiments, and the influence of data noise on TFR is analyzed.
\end{abstract}


\section{Introduction}
Heat monitoring analysis plays an important role in many expensive engineering systems \cite{Yao2011,Zheng2019,Zheng2020}, such as airplanes, satellites, rockets, etc. If the system overheats, it can lead to system failure or even serious disaster. Generally, the heat monitoring analysis is performed using the temperatures measured by many sensors. However, a limited number of temperature monitoring sensors can not provide detailed temperature information at any location of the system, which will result in the heat monitoring analysis results being inaccurate. Therefore, it is essential to reconstruct the temperature field of the whole system based on the temperature measured by sensors.

In recent years, CNN has been applied in many engineering problems \cite{Zhang2020, Zhaoxiaoyu2020, Ronneberger2015} due to its powerful image feature extraction ability. For the TFR, it is a complex image-to-image regression problem. Thus, this paper chooses CNN to reconstruct the temperature field using sensor monitoring temperatures as inputs. It is important to notice that engineers hardly obtain each input's label, i.e., temperature field. However, the training of CNN usually requires a large amount of labeled training data. In order to solve the problem of insufficient labeled data, the physics-informed CNN is used to bearing fault detection \cite{Shen2021}, partial differential equation solution \cite{Fang2021A,Gao2021PhyGeoNet,Gao2021Super}, topology optimization \cite{Zhang2021TONR}. For the TFR, the steady-state temperature field meets the Laplace equation \cite{Gongzq2021,Zhaoxiaoyu2020} and some boundary conditions. Thus, this paper uses physical knowledge to guide the training of CNN.

The training of CNN is essentially extracting model parameters from training data. The trained model is then used to make predictions. Generally, the training data may include data noise which will result in aleatoric uncertainty \cite{Postels2019} in the prediction results. Unlike epistemic uncertainty \cite{Gal2016} can be eliminated by increasing the number of training data, aleatoric uncertainty is inherent to the training data with noise. In recent years, many methods \cite{Kendall2017,Lakshminarayanan2017,Liu2019,Tagasovska2019,Malinin2020} have been studied to quantify aleatoric uncertainty. Given the Gaussian distribution input hypothesis, one common approach of aleatoric uncertainty quantification is that one output of the neural network is used to estimate the conditional variance of the prediction by maximizing likelihood estimation \cite{Kendall2017, Lakshminarayanan2017}. The obvious disadvantage of this approach is that it can only quantify Gaussian random noise. Another approach, simultaneous quantile regression (SQR)\cite{Tagasovska2019}, is to quantify aleatoric uncertainty based on quantile regression \cite{Koenker2004,Takeuchi2006}. In the model training process, this approach randomly samples the quantile level between 0 and 1 for each training data and mini-batch. Compared with the previous approach, SQR can estimate the aleatoric uncertainty in data without Gaussian distribution input assumptions or anything else. Inspired by this approach, this paper proposes a Deep MC-QR method for reconstructing the temperature field and quantifying aleatoric uncertainty caused by data noise.

The main contributions of this paper include the following two points:
\begin{itemize}
	\item By physical knowledge to guide the training of CNN, the proposed Deep MC-QR method can reconstruct an accurate TFR surrogate model without any labeled training data;
	\item By constructing a quantile level image for each monitoring point temperature image, the proposed Deep MC-QR method can quantify aleatoric uncertainty.
\end{itemize}

\section{Related Work}
\subsection{Quantile Regression}
Suppose that there is a stochastic system $ \bm{Y}=f\left(\bm{X}\right) $ with the random input variable $ \bm{X} $. The cumulative distribution function of the output $ \bm{Y} $ taking the value $ \bm{y} $ is $ F_{\bm{Y}|\bm{X}=\bm{x}}\left(\bm{y}|\bm{x}\right)=P\left(\bm{Y} \le \bm{y}\right) $. Thus, the $\tau$ quantile $Q_{\tau}\left(\bm{Y}|\bm{X}=\bm{x}\right)$ of the output $\bm{Y}$ is
\begin{equation}
Q_{\tau}\left(\bm{Y}|\bm{X}=\bm{x}\right) = F_{\bm{Y}|\bm{X}=\bm{x}}^{-1}\left(\tau\right),
\end{equation}
where $\tau$ ($0 \le \tau \le 1$) is the quantile level. The quantile regression constructs a surrogate model $ \hat{\bm{Y}}=\hat{f}_{\tau}(\bm{X}) $ to approximate the stochastic system $ \bm{Y}=f\left(\bm{X}\right) $. Given $\mathcal{K}$ training data $\left\{\left(\bm{x}_k, \bm{y}_k\right)| k=1,2,\cdots, \mathcal{K}\right\}$, the surrogate model $ \hat{\bm{Y}}=\hat{f}_{\tau}(\bm{X}) $ is trained by minimizing the pinball loss function $ \mathcal{L}\left[ {{w}_{i}},{{{\hat{f}}}_{\tau}}\left( \bm{v}_i \right) \right] $ \cite{Takeuchi2006, Tagasovska2019}, i.e., 
\begin{equation}
\setlength{\abovedisplayskip}{3pt}
\setlength{\belowdisplayskip}{3pt}
\begin{aligned}
& \mathcal{L}\left[ {{\bm{y}}_{k}},{{{\hat{f}}}_{\tau}}\left( \bm{x}_k \right) \right]=\frac{1}{\mathcal{K}}\sum\limits_{k=1}^{\mathcal{K}}{{{\ell}_{\tau}}\left[ {{\bm{y}}_{k}},{{{\hat{f}}}_{\tau}}\left( {{\bm{x}}_{k}} \right) \right]}, \\
& {{\ell}_{\tau}}\left[ {{\bm{y}}_{k}},{{{\hat{f}}}_{\tau}}\left( {{\bm{x}}_{k}} \right) \right]=\left\{ \begin{matrix}
\tau\left( {{\bm{y}}_{k}}-{{{\hat{\bm{y}}}}_{k}} \right), & {{\bm{y}}_{k}}\ge {{{\hat{\bm{y}}}}_{k}}  \\
\left( 1-\tau \right)\left( {{{\hat{\bm{y}}}}_{k}}-{{\bm{y}}_{k}} \right), & {{\bm{y}}_{k}}<{{{\hat{\bm{y}}}}_{k}}  \\
\end{matrix} \right.,
\end{aligned}
\end{equation}
where ${{\bm{\hat{y}}}_{k}}={{\hat{f}}_{\tau}}\left( {{\bm{x}}_{k}} \right)$ is the prediction corresponding to the input $ \bm{x}_{k} $.

\subsection{Temperature Field Reconstruction}
\begin{figure}[htbp]
	\setlength{\abovecaptionskip}{-0.03cm}
	\centering
	{\includegraphics[scale=0.7]{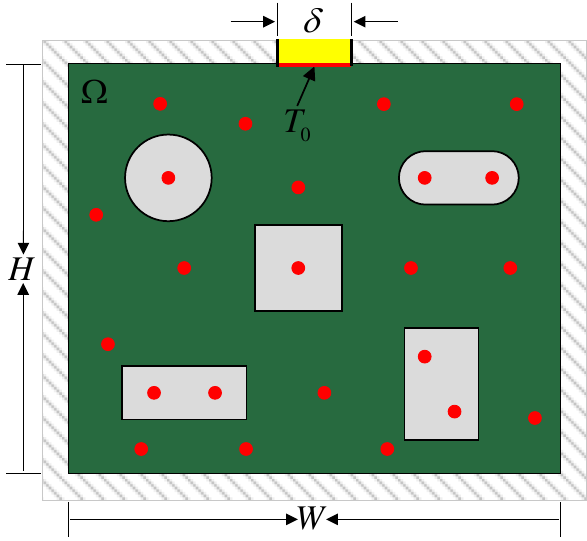}}
	\caption{A rectangular area heat source system.}
	\label{TFR}
\end{figure}

As shown in Fig.\ref{TFR}, there is an green rectangular area $\Omega$ with length $ W $ and width $ H $, where the grey areas $ {\Omega}_{HS} $ of circles, capsules and rectangles are heat sources. The yellow rectangle with width $ \delta $ is the position of the heat sink. The diagonal shaded area around $\Omega$ is adiabatic. Besides, the red points are the positions $ {\Omega}_{MP} $ of the temperature monitoring sensors. Supposed that there are $ N_{MP} $ temperature monitoring sensors in the rectangular area $\Omega$.

For this green rectangular area $\Omega$, its steady-state temperature field $ \bm{T} $ meets the Laplace equation \cite{Zhaoxiaoyu2020, Gongzq2021}, i.e.,
\begin{equation}\label{laplace}
\setlength{\abovedisplayskip}{3pt}
\setlength{\belowdisplayskip}{3pt}
\frac{{{\partial }^{2}}\bm{T}\left( u,v \right)}{\partial {{u}^{2}}}+\frac{{{\partial }^{2}}\bm{T}\left( u,v \right)}{\partial {{v}^{2}}}+\bm{\varphi }\left( u,v \right)=0, \quad \left( u,v \right)\in \Omega,
\end{equation}
and the boundary condition
\begin{equation}\label{BC}
\setlength{\abovedisplayskip}{3pt}
\setlength{\belowdisplayskip}{3pt}
\bm{T}\left( u,v \right)={{T}_{0}},\quad \left( u,v \right)\in {{\Omega }_{BC}},
\end{equation}
where $\left( u,v \right)$ denotes the position coordinate, $T_0$ is a constant temperature, $\bm{\varphi }$ is the heat source intensity distribution depending on the power of heat source, and $ {{\Omega }_{BC}} $ denotes the boundary (red line in Fig.\ref{TFR}). Generally, the heat source intensity distribution $\bm{\varphi }$ is unknown in the TFR problem. Thus, the TFR problem only considers the heat conduction in the area ${\Omega }_{NC}$ without heat sources, i.e.,
\begin{equation}\label{laplace_nmp}
\setlength{\abovedisplayskip}{3pt}
\setlength{\belowdisplayskip}{3pt}
\frac{{{\partial }^{2}}\bm{T}\left( u,v \right)}{\partial {{u}^{2}}}+\frac{{{\partial }^{2}}\bm{T}\left( u,v \right)}{\partial {{v}^{2}}}=0, \quad \left( u,v \right)\in {{\Omega }_{NC}}.
\end{equation}

\begin{figure}[htbp]
	\vspace{-0.2cm} 
	\setlength{\abovecaptionskip}{-0.005cm} 
	\centering
	{\includegraphics[scale=0.65]{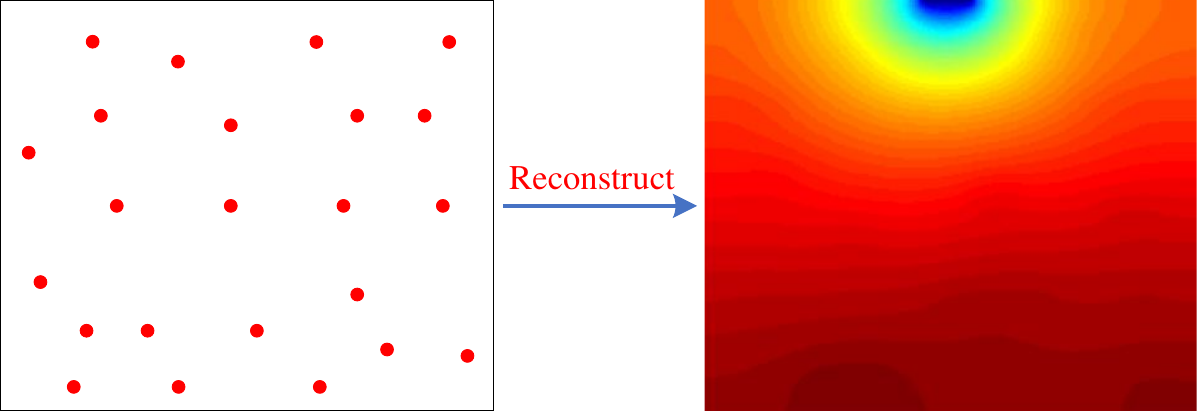}}
	\caption{Temperature Field Reconstruction by monitoring point temperatures.}
	\label{TFR_problem}
\end{figure}

For the TFR problem, its objective is to reconstruct the temperature field of the rectangular area $\Omega$ from the temperatures of some monitoring points, as shown in Fig.\ref{TFR_problem}. 

\section{Deep Monte Carlo Quantile Regression}
\subsection{Physics-informed Deep MC-QR model}\label{sect31}
\begin{figure}[htbp]
	\vspace{-0.26cm} 
	\setlength{\abovecaptionskip}{-0.03cm} 
	\centering
	{\includegraphics[scale=0.65]{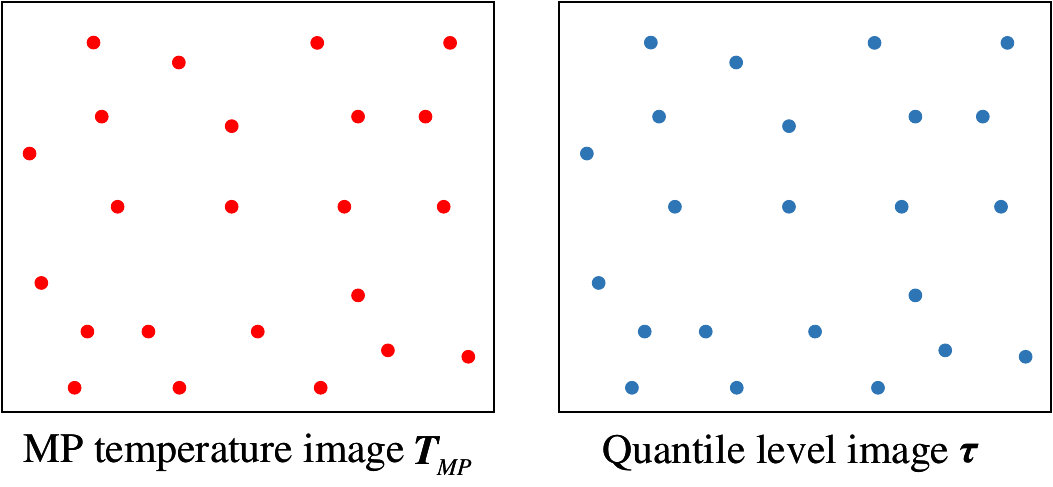}}
	\caption{Temperature field reconstruction by monitoring point temperatures.}
	\label{CNN_input}
\end{figure}

In this paper, the rectangular area $\Omega$ is discretized into a $ h \times w $ two-dimensional array. According to the positions $ {\Omega}_{MP} $ of the temperature monitoring sensors, the corresponding element values in the $ h \times w $ two-dimensional array are equal to the the monitoring point temperatures. The rest of element values are set to be zero. Thereby, a monitoring point (MP) temperature image $ \bm{T}_{MP} $ is constructed as shown in Fig.\ref{CNN_input} (left). Besides, this paper also builds a quantile level image $ \bm{\tau} $ as shown in Fig.\ref{CNN_input} (right). Different from the MP temperature image $ \bm{T}_{MP} $, the values of element in the positions $ {\Omega}_{MP} $ are equal to $ \tau $ for the quantile level image $ \bm{\tau} $, where $ \tau $ is randomly sampled from the uniform distribution $ U\left(0, 1\right) $.

\begin{figure}[!htbp]
	\setlength{\abovecaptionskip}{-0.03cm}
	\centering
	{\includegraphics[scale=0.35]{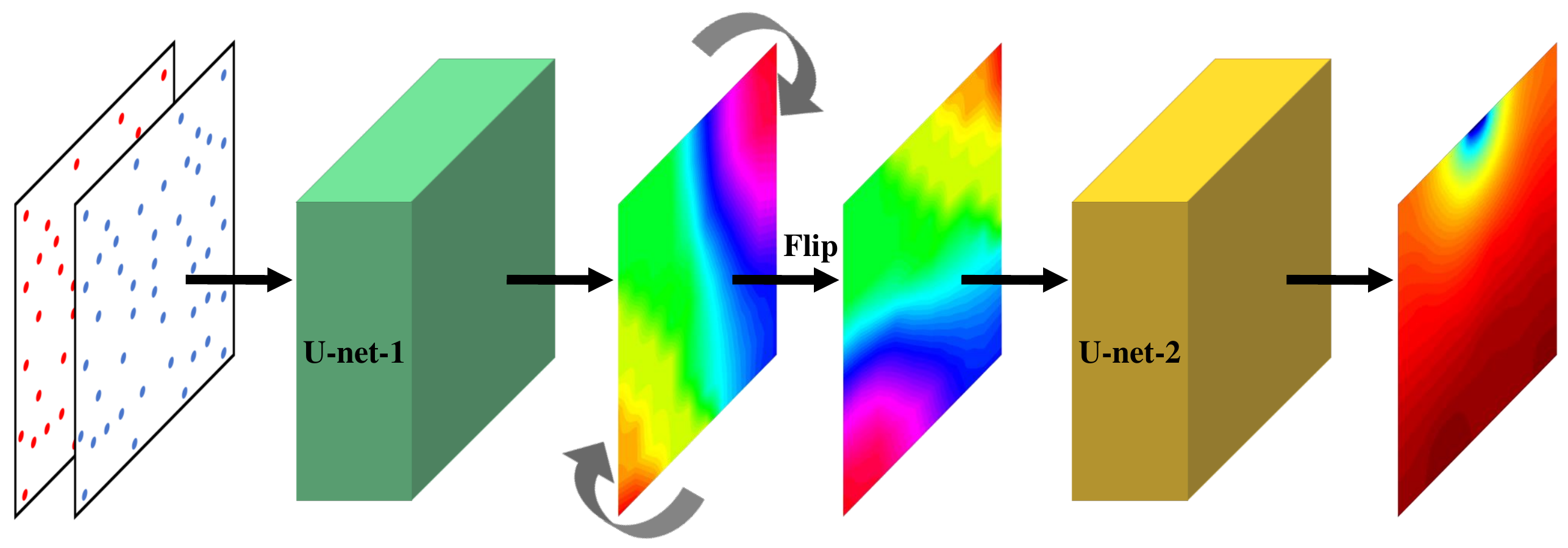}}
	\caption{The physics-informed Deep MC-QR model.}
	\label{MCQR_CNN}
\end{figure}

This paper builds a physics-informed Deep MC-QR model which can use the MP temperature image $ \bm{T}_{MP} $ and the quantile level image $ \bm{\tau} $ to reconstruct the temperature field of the rectangular area $\Omega$, as shown in Fig.\ref{MCQR_CNN}. The physics-informed Deep MC-QR model with two input channels includes two U-net \cite{Ronneberger2015} models, i.e., U-net-1 and U-net-2. Firstly, U-net-1 extracts the feature map of input $\left( {{\bm{T}}_{MP}},\bm{\tau } \right)$. Then, the feature map is flipped diagonally to ensure the reconstruction accuracy of temperature field in the upside and right-side boundaries \cite{Gongzq2021}. Finally, U-net-2 uses the flipped feature map to predict the temperature field $\hat{\bm{T}}$ of the rectangular area $\Omega$.

Supposed that the physics-informed Deep MC-QR model is denoted as $ \mathcal{MQNN}\left(\bm{T}_{MP}, \bm{\tau}; \bm{\theta}\right) $, where $ \bm{\theta} $ denotes the parameters of U-net-1 and U-net-2. Based on the training dataset $\left\{\left(\bm{T}^i_{MP}, \bm{\tau}_i\right)|i=1,2,\cdots,N \right\}$, the parameters $\bm{\theta}$ are learned by minimizing the following proposed physics-informed loss function $\mathcal{L}\left(\theta\right)$.
\begin{itemize}
	\item \textbf{Quantile MP temperature loss $ \mathcal{L}_{\tau}\left( \bm{\theta} \right) $}
\end{itemize}

For the $i\text{th}$ MP temperature image $ \bm{T}^{i}_{MP} $, the predicted temperature field $\hat{\bm{T}}$ of the satellite subsystem is $ {\hat{\bm{T}}_i}=\mathcal{M}\left(\bm{T}^{i}_{MP}, \bm{\tau}_{i}; \bm{\theta}\right) $. For $ \left ( u,v \right ) \in {\Omega}_{MP} $, the error $ {{\ell}_{{\tau}_i}}\left( u,v;\bm{\theta} \right) $ is
\begin{equation}
\setlength{\abovedisplayskip}{3pt}
\setlength{\belowdisplayskip}{3pt}
{{\ell}_{{\tau}_i}}\left( u,v;\bm{\theta} \right) = {\tau}_i\left[ {{\bm{T}^i_{MP}}\left(u,v\right)}-{{\hat{\bm{T}_i}}\left( u,v \right)} \right]
\end{equation}
for the condition $ {{\bm{T}^i_{MP}}\left(u,v\right)}\ge {{\hat{\bm{T}_i}}\left( u,v \right)} $, and
\begin{equation}
\setlength{\abovedisplayskip}{3pt}
\setlength{\belowdisplayskip}{3pt}
{{\ell}_{{\tau}_i}}\left( u,v;\bm{\theta} \right) = \left( 1-{\tau}_i\right)\left( {{\hat{\bm{T}_i}}\left( u,v \right)}-{{\bm{T}^i_{MP}}\left(u,v\right)} \right)
\end{equation}
for the condition $ {{\bm{T}^i_{MP}}\left(u,v\right)}<{{\hat{\bm{T}_i}}\left( u,v \right)} $, where ${\tau}_i \sim U\left ( 0, 1 \right ) $. Thus, the quantile MP temperature loss $ \mathcal{L}_{\tau}\left( \bm{\theta} \right) $ is
\begin{equation}\label{QMPT_loss}
\setlength{\abovedisplayskip}{3pt}
\setlength{\belowdisplayskip}{3pt}
\mathcal{L}_{\tau}\left( \bm{\theta} \right)=\frac{1}{N}\frac{1}{N_{MP}}\sum\limits_{i=1}^{N}{\sum_{\left ( u,v \right ) \in {\Omega}_{MP}}  {{{\ell}_{{\tau}_i}}\left( u,v;\bm{\theta} \right)}}.
\end{equation}

\begin{itemize}
	\item \textbf{Laplace equation loss $ \mathcal{L}_{LE}\left( \bm{\theta} \right) $}
\end{itemize}

Assumed that each discretized MP temperature image $ \bm{T}^{i}_{MP} $ has $N_{NC}$ points for $ \left( u,v \right)\in {{\Omega }_{NC}} $. Discretizing the Laplace equation (\ref{laplace_nmp}), ${{{\partial }^{2}}\hat{\bm{T}}_i\left( u_j,v_k \right)}/{\partial {{u}_j^{2}}}$ and $ {{\partial }^{2}}\hat{\bm{T}}_i\left( u_j,v_k \right)/{\partial {{v}_k^{2}}} $ are transformed respectively to be
\begin{equation}\label{le_X_discre_simple}
\setlength{\abovedisplayskip}{3pt}
\setlength{\belowdisplayskip}{3pt}
\frac{{{\partial }^{2}}\hat{\bm{T}}_i\left( u_j,v_k \right)}{\partial {{u}_j^{2}}}=\frac{\Delta {{{\hat{\bm{T}}}}^i_{j+1}}\Delta {{u}_{j}}-\Delta {{{\hat{\bm{T}}}}^i_{j}}\Delta {{u}_{j+1}}}{\Delta {{u}_{j+1}}{{\Delta }^{2}}{{u}_{j}}}
\end{equation}
and
\begin{equation}\label{le_Y_discre_simple}
\setlength{\abovedisplayskip}{3pt}
\setlength{\belowdisplayskip}{3pt}
\frac{{{\partial }^{2}}\hat{\bm{T}}_i\left( {{u}_{j}},{{v}_{k}} \right)}{v_{k}^{2}}=\frac{\Delta {{{\hat{\bm{T}}}}^i_{k+1}}\Delta {{v}_{k}}-\Delta {{{\hat{\bm{T}}}}^i_{k}}\Delta {{v}_{k+1}}}{\Delta {{v}_{k+1}}{{\Delta }^{2}}{{v}_{k}}}, 
\end{equation}
where $ \Delta {{u}_{j}}={{u}_{j}}-{{u}_{j-1}} $, $ \Delta {{v}_{k}}={{v}_{k}}-{{v}_{k-1}} $, $ \Delta {{\hat{\bm{T}}}^i_{j}}=\hat{\bm{T}}_i\left( {{u}_{j}},{{v}_{k}} \right)-\hat{\bm{T}}_i\left( {{u}_{j-1}},{{v}_{k}} \right) $ and $ \Delta {{\hat{\bm{T}}}^i_{k}}=\hat{\bm{T}}_i\left( {{u}_{j}},{{v}_{k}} \right)-\hat{\bm{T}}_i\left( {{u}_{j}},{{v}_{k-1}} \right) $. Thus, the Laplace equation loss $ \mathcal{L}_{LE}\left( \bm{\theta} \right) $ for $ {\left( u,v \right)\in {{\Omega }_{NC}}} $ is
\begin{equation}\label{LE_loss}
\setlength{\abovedisplayskip}{3pt}
\setlength{\belowdisplayskip}{3pt}
{{\mathcal{L}}_{LE}}\left( \bm{\theta } \right)=\frac{1}{N}\frac{1}{{{N}_{NC}}}\sum\limits_{i=1}^{N}{\sum\limits_{\left( u,v \right)\in {{\Omega }_{NC}}}{{{\left| \frac{{{\partial }^{2}}\hat{\bm{T}}_i\left( u,v \right)}{\partial {{u}^{2}}}+\frac{{{\partial }^{2}}\hat{\bm{T}}_i\left( u,v \right)}{\partial {{v}^{2}}} \right|}^{2}}}}.
\end{equation}

\begin{itemize}
	\item \textbf{Boundary condition loss $ \mathcal{L}_{BC}\left( \bm{\theta} \right) $}
\end{itemize}

According to (\ref{BC}), the boundary condition loss $ \mathcal{L}_{BC}\left( \bm{\theta} \right) $ for $N$ MP temperature images $\left\{ \bm{T}^i_{MP}|i=1,2,\cdots,N \right\}$ is
\begin{equation}\label{bc_loss}
\setlength{\abovedisplayskip}{3pt}
\setlength{\belowdisplayskip}{3pt}
{{\mathcal{L}}_{BC}}\left( \bm{\theta } \right)=\frac{1}{N}\frac{1}{{{N}_{BC}}}\sum\limits_{i=1}^{N}{\sum\limits_{\left( u,v \right)\in {{\Omega }_{BC}}}{{{\left| {{{\hat{\bm{T}}}}_{i}}\left( u,v \right)-{{T}_{0}} \right|}^{2}}}}.
\end{equation}
where $N_{BC}$ means the number of points $\left(u,v\right)$ in the area ${{\Omega }_{BC}}$ of each discretized  MP temperature image $ \bm{T}^{i}_{MP} $.

\begin{itemize}
	\item \textbf{TV regularization $ \mathcal{L}_{TV}\left( \bm{\theta} \right) $}
\end{itemize}

TV regularization \cite{Rudin1992, Gongzq2021}, as shown in (\ref{tv}), can maintain the smoothness of the image, 
\begin{equation}\label{tv}
\setlength{\abovedisplayskip}{3pt}
\setlength{\belowdisplayskip}{3pt}
{{\mathcal{R}}_{{{V}^{\beta }}}}=\int_{\Omega }{{{\left\{ {{\left[ \frac{\partial f\left( u,v \right)}{\partial u} \right]}^{2}}+{{\left[ \frac{\partial f\left( u,v \right)}{\partial v} \right]}^{2}} \right\}}^{\frac{\beta }{2}}}}dudv,
\end{equation}
where $ f\left(u,v\right) $ is a continuous function. Due to the property that the steady-state temperature field will not mutate sharply, this paper adopts TV regularization to assist the training of model $ \mathcal{MQNN}\left(\bm{T}_{MP}, \bm{\tau}; \bm{\theta}\right) $. The TV regularization $ \mathcal{L}_{TV}\left( \bm{\theta} \right) $ ($\beta=2$) is transformed to be (\ref{tv_loss}) by the finite-difference approximation, i.e.,
\begin{equation}\label{tv_loss}
\setlength{\abovedisplayskip}{3pt}
\setlength{\belowdisplayskip}{3pt}
\begin{aligned}
& \mathcal{L}_{TV}^{u}\left( \bm{\theta } \right)=\frac{1}{h\times \left(w-1\right)}\sum\limits_{k=1}^{h}{\sum\limits_{j=1}^{w-1}{{{\left( \Delta\hat{\bm{T}}_{j}^{TV} \right)}^{2}}}}, \\ 
& \mathcal{L}_{TV}^{v}\left( \bm{\theta } \right)=\frac{1}{\left( h-1 \right)\times w}\sum\limits_{j=1}^{w}{\sum\limits_{k=1}^{h-1}{{{\left( \Delta\hat{\bm{T}}_{k}^{TV} \right)}^{2}}}}, \\ 
& {{\mathcal{L}}_{TV}}\left( \bm{\theta } \right)=\frac{1}{N}\sum\limits_{i=1}^{N}{\left[ \mathcal{L}_{TV}^{u}\left( \bm{\theta } \right)+\mathcal{L}_{TV}^{v}\left( \bm{\theta } \right) \right]}, \\ 
\end{aligned}
\end{equation}
where $ \Delta\hat{\bm{T}}_{j}^{TV} = {{{\bm{\hat{T}}}}_{i}}\left( {{u}_{j+1}},{{v}_{k}} \right)-{{{\bm{\hat{T}}}}_{i}}\left( {{u}_{j}},{{v}_{k}} \right) $ and $ \Delta\hat{\bm{T}}_{k}^{TV} = {{{\bm{\hat{T}}}}_{i}}\left( {{u}_{j}},{{v}_{k+1}} \right)-{{{\bm{\hat{T}}}}_{i}}\left( {{u}_{j}},{{v}_{k}} \right) $

Thus, the proposed physics-informed loss function $\mathcal{L}\left(\theta\right)$ for learning the parameters $\bm{\theta}$ of model $\mathcal{MQNN}\left(\bm{T}_{MP}, \bm{\tau}; \bm{\theta}\right)$ is
\begin{equation}\label{PI_Loss}
\setlength{\abovedisplayskip}{3pt}
\setlength{\belowdisplayskip}{3pt}
\begin{aligned}
& \mathcal{L}\left( \bm{\theta } \right)={\alpha }_{1}{{\mathcal{L}}_{\tau }}\left( \bm{\theta } \right)+{{\alpha }_{2}}{{\mathcal{L}}_{LE}}\left( \bm{\theta } \right)+{{\alpha }_{3}}{{\mathcal{L}}_{BC}}\left( \bm{\theta } \right) \\
& \qquad \quad\; +{{\alpha }_{4}}{{\mathcal{L}}_{TV}}\left( \bm{\theta } \right),
\end{aligned}
\end{equation}
where $ {\alpha }_{1} $, $ {\alpha }_{2} $, $ {\alpha }_{3} $, and $ {\alpha }_{4} $ are hyperparameters.

\subsection{Model Monte Carlo Training Algorithm}
This section proposes the model Monte Carlo training algorithm to learning the parameters $\bm{\theta}$. Different from the general model training methods, the model Monte Carlo training algorithm keeps changing the training data by randomly sampling quantile level images in each epoch. The detailed description is as follows: In the $ep\text{-th}$ epoch, a quantile level image $ \bm{\tau}_i $ ($ i=1,2,\cdots,N $) corresponding to the MP temperature image $ \bm{T}^i_{MP} $ is created by the method in section \ref{sect31}. Then, the quantile level image $ \bm{\tau}_i $ and the MP temperature image $ \bm{T}^i_{MP} $ ($ i=1,2,\cdots,N $) constitute the training dataset $\left\{\left(\bm{T}^i_{MP}, \bm{\tau}_i\right)_{ep}|i=1,2,\cdots,N \right\}$. Set the appropriate batch size, and the training dataset is divided into $N_{batch}$ batches. This paper chooses the Adam algorithm \cite{Kingma2014} to update the parameters $\bm{\theta}$. In summary, the pseudo code of the model Monte Carlo training algorithm is shown in \textbf{Algorithm} \ref{algorithm1}.

\begin{algorithm}[!htbp]
	\caption{Model Monte Carlo Training Algorithm.}
	\label{algorithm1}
	\LinesNumbered
	\KwIn{\\ \qquad MP temperature images $\left\{ \bm{T}^i_{MP}|i=1,2,\cdots,N \right\}$.
	}
	\KwOut{\\ \qquad Trained model $\mathcal{MQNN}\left(\bm{T}_{MP}, \bm{\tau}; \bm{\theta}\right)$.}
	Initialize the physics-informed Deep MC-QR model $\mathcal{MQNN}\left(\bm{T}_{MP}, \bm{\tau}; \bm{\theta}\right)$; \\ 
	\For{$ep=1:ep_{max}$}{
		\For{$i=1:N$}{
			Randomly take a sample ${\tau}_i$ from uniform distribution $U\left ( 0, 1 \right )$; \\
			Create a ${h}\times {w}$ two-dimensional array $\bm{\tau}_i$ with all zero elements; \\
			Perform $ \bm{\tau}_i\left(u,v\right) = {\tau}_i$ for $ \forall \left( u,v \right) \in {{\Omega}_{MP}} $; \\
			Composite training data $ \left( \bm{T}_{MP}^{i},{{\bm{\tau }}_{i}} \right)_{ep} $.
		}
		Dividing $N$ training data $\left\{\left(\bm{T}^i_{MP}, \bm{\tau}_i\right) \right\}$ to be $ N_{batch} $ batches; \\
		\For{$ba=1:N_{batch}$}{
			Estimate the reconstructed temperature field ${{\hat{\bm{T}}}_{i}}=\mathcal{MQNN}\left(\bm{T}^i_{MP}, \bm{\tau}_i; \bm{\theta}\right)$;\\
			Calculate the proposed loss function $\mathcal{L}\left( \bm{\theta } \right)$; \\
			Calculate the gradient of $\mathcal{L}\left( \bm{\theta } \right)$; \\
			Update the parameters $ \bm{\theta} $ by the Adam algorithm.
	    }
	}
\end{algorithm}

\subsection{Model Prediction and Aleatoric Uncertainty Quantification}
For a MP temperature image $\bm{T}^{pre}_{MP}$, the reconstructed temperature field $ {{\hat{\bm{T}}}_{pre}} $ and the quantified aleatoric uncertainty $ \hat{\bm{\sigma}}_{pre} $ are calculated as follows. In the predicted stage, $N_{pre}$ quantile level images are generated by the method in section \ref{sect31}. Then, the composited prediction data $ \left\{ \left(\bm{T}^{pre}_{MP},\bm{\tau}_{\mathcal{A} }^{pre}\right)|\mathcal{A}=1,2,\cdots,N_{pre} \right\} $ are obtained for the MP temperature image $\bm{T}^{pre}_{MP}$. For $ \mathcal{A}=1,2,\cdots,N_{pre} $, the temperature field $ \hat{\bm{T}}_{\mathcal{A} }^{pre} $ is predicted by the trained model $\mathcal{MQNN}\left(\bm{T}_{MP}, \bm{\tau}; \bm{\theta}\right)$. Thereby, the reconstructed temperature field's approximation $\hat{\bm{T}}_{pre}$ and the aleatoric uncertainty $\bm{\sigma}_{pre}$ are estimated by calculating the mean and the standard deviation of $ N_{pre} $ results $ \hat{\bm{T}}_{\mathcal{A} }^{pre} $ ($\mathcal{A}=1,2,\cdots,N_{pre}$), respectively, i.e.,
\begin{equation}\label{T_pre}
\setlength{\abovedisplayskip}{3pt}
\setlength{\belowdisplayskip}{1pt}
{{\hat{\bm{T}}}_{pre}}=\frac{1}{N_{pre}}\sum\limits_{\mathcal{A}=1}^{{{N}_{pre}}}{\hat{\bm{T}}_{\mathcal{A}}^{pre}},
\end{equation}

\begin{equation}\label{U_alea}
\setlength{\abovedisplayskip}{1pt}
\setlength{\belowdisplayskip}{3pt}
{{\bm{\sigma }}_{pre}}={{\left[ \frac{1}{N_{pre}}\sum\limits_{\mathcal{A}=1}^{{{N}_{pre}}}{{{\left( \hat{\bm{T}}_{\mathcal{A}}^{pre} \right)}^{2}}-{{\left( {{{\hat{\bm{T}}}}_{pre}} \right)}^{2}}} \right]}^{\frac{1}{2}}}.
\end{equation}
In summary, the pseudo code of the model prediction and aleatoric uncertainty quantification algorithm is shown in \textbf{Algorithm} \ref{algorithm2}.

\begin{algorithm}[!htbp]
	\caption{Model prediction and aleatoric uncertainty quantification algorithm.}
	\label{algorithm2}
	\LinesNumbered
	\KwIn{\\ 
		\qquad (1) MP temperature image $\bm{T}^{pre}_{MP}$; \\
		\qquad (2) Sampling number $ N_{pre} $ of quantile levels; \\
		\qquad (3) Trained model $\mathcal{MQNN}\left(\bm{T}_{MP}, \bm{\tau}; \bm{\theta}\right)$. \\
	}
	\KwOut{\\ 
		\qquad (1) Reconstructed temperature field $ {{\hat{\bm{T}}}_{pre}} $; \\
		\qquad (2) Quantified aleatoric uncertainty $ \hat{\bm{\sigma}}_{pre} $. \\
	}
	
	\For{$\mathcal{A}=1:N_{pre}$}{
		Randomly take a sample ${\tau}^{pre}_{\mathcal{A}}$ from uniform distribution $U\left ( 0, 1 \right )$; \\
		Create a $h\times w$ two-dimensional array $\bm{\tau}^{pre}_{\mathcal{A}}$ with all zero elements; \\
		Perform $\bm{\tau}^{pre}_{\mathcal{A}}\left(u,v\right) = {\tau}^{pre}_{\mathcal{A}}$ for $ \forall \left( u,v \right) \in {{\Omega}_{MP}} $; \\
		Composite prediction data $ \left( \bm{T}^{pre}_{MP},\bm{\tau}^{pre}_{\mathcal{A}} \right) $;\\
		Predict the reconstructed temperature field ${{\hat{\bm{T}}}^{pre}_{\mathcal{A}}}=\mathcal{M}\left(\bm{T}^{pre}_{MP},\bm{\tau}^{pre}_{\mathcal{A}}; \bm{\theta}\right)$.
	}
	Calculate the mean $ {{\hat{\bm{T}}}_{pre}} $ and the standard deviation $ \hat{\bm{\sigma}}_{pre} $ of $ \left\{{{\hat{\bm{T}}}^{pre}_{\mathcal{A}}}|\mathcal{A}=1,2,\cdots,N_{pre}\right\} $. \\
\end{algorithm}

\section{Experiment}
\subsection{Experimental Setup}\label{sec41}
\begin{figure}[!htbp]
	\setlength{\belowcaptionskip}{-0.3cm}
	\centering
	{\includegraphics[scale=0.4]{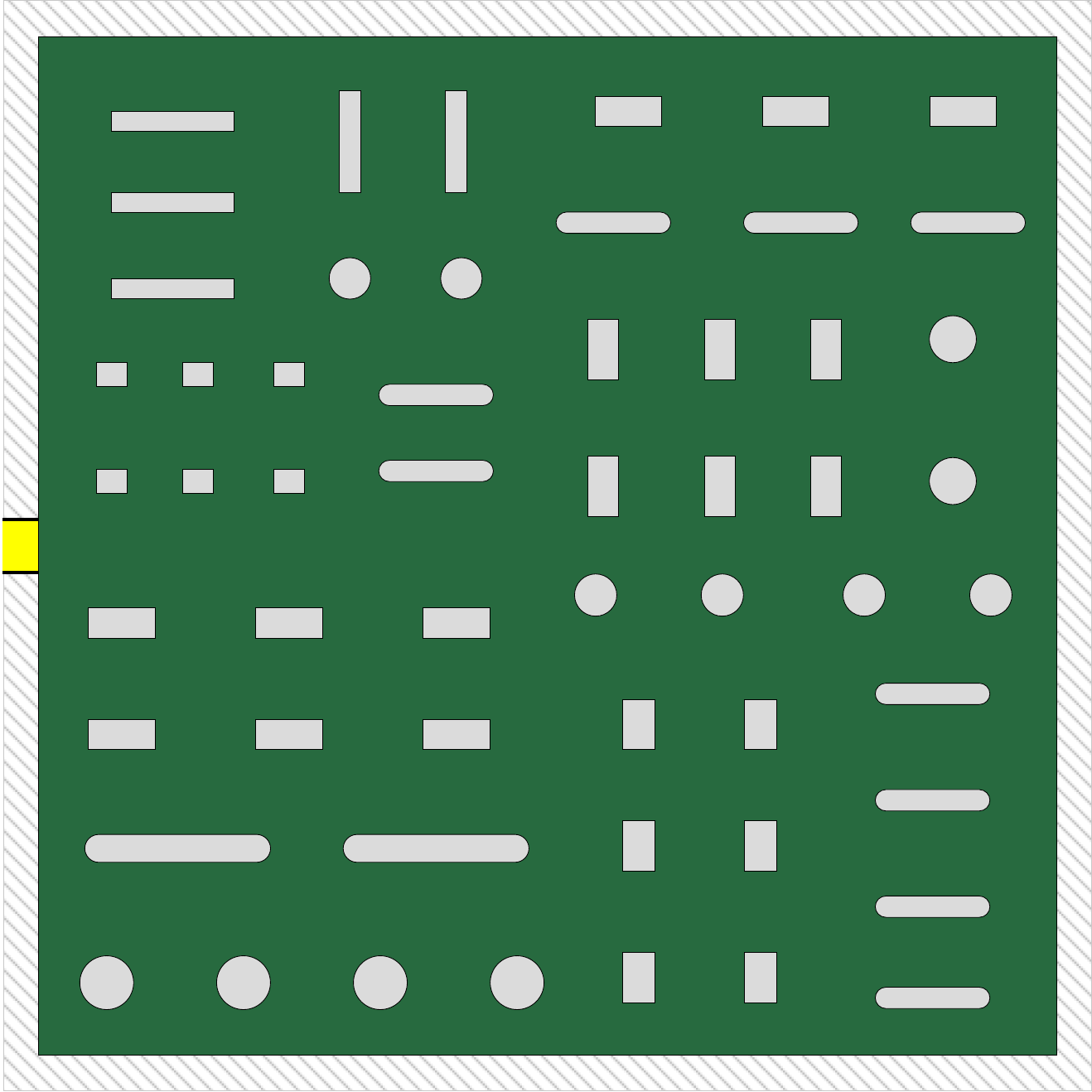}}
	\caption{Heat source layout.}
	\label{layout_case}
\end{figure}

\begin{figure}[!htbp]
	\setlength{\belowcaptionskip}{-0.3cm}
	\centering
	{\includegraphics[scale=0.4]{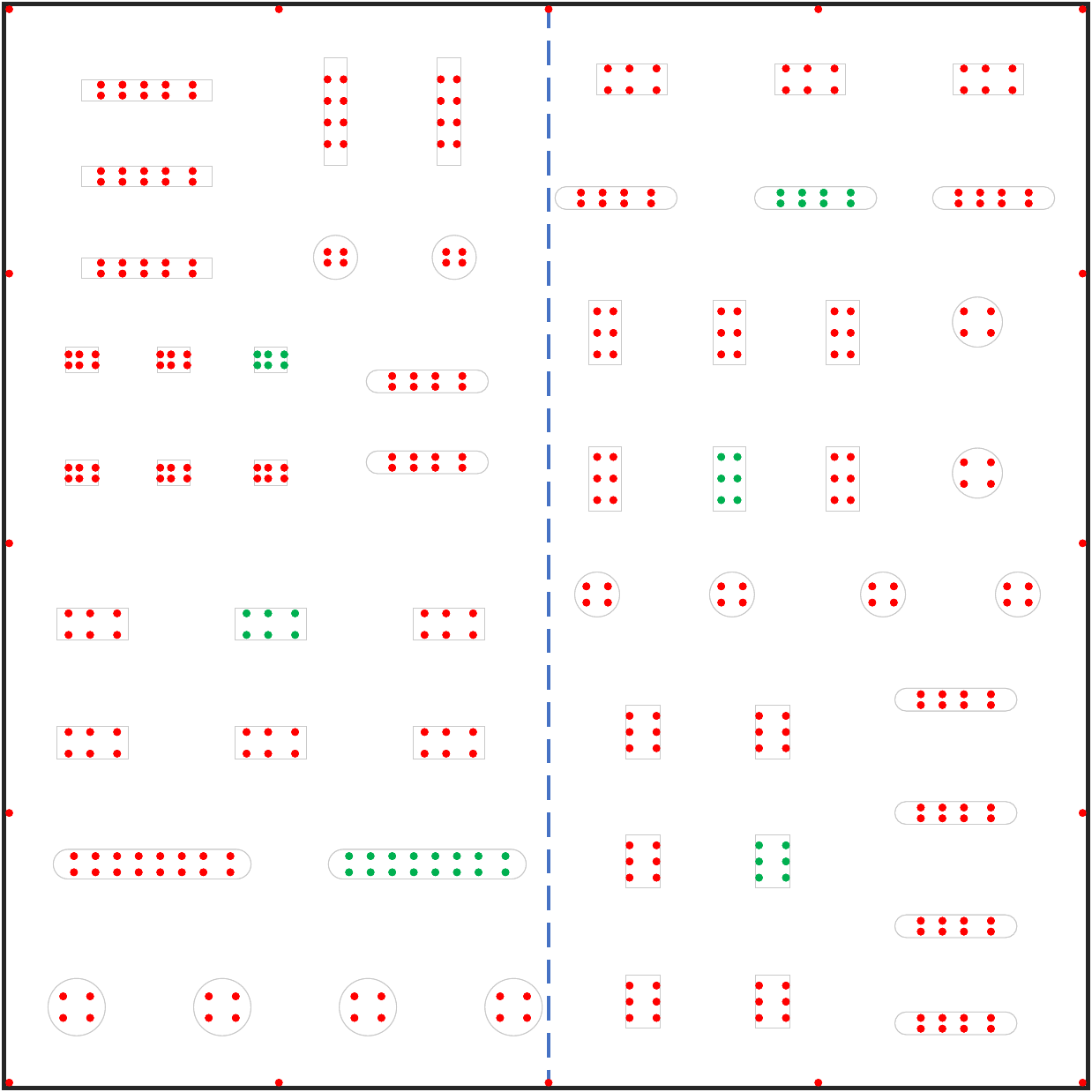}}
	\caption{Monitoring point positions.}
	\label{MP_case}
\end{figure}

As shown in Fig.\ref{layout_case}, this paper uses a system ($W=H=0.2 \;\text{m}$) to validate the effectiveness of the proposed Deep MC-QR method. This system has 55 heat sources, and the heat sink with width $ \delta=0.02\;\text{m} $ is in the center of the left boundary of the system. The positions of monitoring points are shown in Fig.\ref{MP_case}. In this experiment, the power of each heat source obeys the normal distribution with a mean of 20000 W and a standard deviation of 1000 W. Besides, the constant temperature $ T_0 $ is 298 K. The parameters $ h $ and $ w $ are set to be 200, i.e., $ h=w=200 $. 15000 MP temperature images $\left\{{\bm{T}}_{MP}^1, {\bm{T}}_{MP}^2, \cdots ,{\bm{T}}_{MP}^{15000}\right\}$ are generated by the recon-data-generator\footnote{https://github.com/shendu-sw/recon-data-generator}, where the training dataset, the validating dataset, and the testing dataset have 10800, 1200 and 3000 MP temperature images, respectively. Especially, the corresponding truth temperature fields (label) of 3000 testing data are also generated by the recon-data-generator. The hyperparameters $ {\alpha }_1 $, $ {\alpha }_2 $, $ {\alpha }_3 $, and $ {\alpha }_4 $ are equal to $1\times10^5$, $1\times10^2$, $1\times10^2$ and $1\times10^4$, respectively. The relevant codes of the Deep MC-QR method by Python are available on this website \footnote{https://github.com/Xiaohu-Zheng/Deep-MC-QR}.

In this experiment, five kinds of noises are respectively added into temperature values of 15000 MP temperature images $\left\{{\bm{T}}_{MP}^1, {\bm{T}}_{MP}^2, \cdots ,{\bm{T}}_{MP}^{15000}\right\}$ to generate five types of the dataset as follows:
\begin{itemize}
	\item Gaussian noise $ {\varepsilon}_1 \sim N\left ( 0, {0.3}^2  \right )  $ for green monitoring points in Fig.\ref{MP_case}.
	\item Gaussian noise $ {\varepsilon}_2 \sim N\left ( 0, {0.5}^2  \right )  $ for green monitoring points in Fig.\ref{MP_case}.
	\item Uniform noise $ {\varepsilon}_3 \sim U\left ( -0.3, 0.3  \right )  $ for green monitoring points in Fig.\ref{MP_case}.
	\item Uniform noise $ {\varepsilon}_4 \sim U\left ( -1, 1  \right )  $ for green monitoring points in Fig.\ref{MP_case}.
	\item Gaussian noise $ {\varepsilon}_5 \sim N\left ( 0, {0.3}^2  \right )  $ for monitoring points to the right of the blue dotted line in Fig.\ref{MP_case} (Excluding the five monitoring points on the right boundary).
\end{itemize}
Based on the above five kinds of the dataset, five trained models are denoted as $\mathcal{MQNN}_{{\varepsilon}_1}$, $\mathcal{MQNN}_{{\varepsilon}_2}$, $\mathcal{MQNN}_{{\varepsilon}_3}$, $\mathcal{MQNN}_{{\varepsilon}_4}$, and $\mathcal{MQNN}_{{\varepsilon}_5}$, respectively. Besides, four evaluation criteria for model accuracy are as follows:
a) the average of root mean square error (RMSE)
\begin{equation}\label{e_RMSE}
\setlength{\abovedisplayskip}{3pt}
\setlength{\belowdisplayskip}{3pt}
\overline{RMSE}=\frac{1}{6\text{e}5}\sum\limits_{t=1}^{3000}{\sqrt{\sum\limits_{\left( x,y \right)\in \Omega }{{{\left[ \bm{T}_{t}\left( x,y \right)-\bm{\hat{T}}_{t}\left( x,y \right) \right]}^{2}}}}},
\end{equation}	
b) the average of mean absolute error (MAE)
\begin{equation}\label{e_MAE}
\setlength{\abovedisplayskip}{3pt}
\setlength{\belowdisplayskip}{3pt}
\overline{MAE}=\frac{1}{1.2\text{e}8}\sum\limits_{t=1}^{3000}{\sum\limits_{\left( x,y \right)\in \Omega }{\left| \bm{T}_{t}\left( x,y \right)-\bm{\hat{T}}_{t}\left( x,y \right) \right|}},
\end{equation}
c) the average of mean relative error (MRE)
\begin{equation}\label{e_MRE}
\setlength{\abovedisplayskip}{3pt}
\setlength{\belowdisplayskip}{3pt}
\overline{MRE}=\frac{1}{1.2\text{e}8}\sum\limits_{t=1}^{3000}{\sum\limits_{\left( x,y \right)\in \Omega }{\frac{\left| \bm{T}_{t}\left( x,y \right)-\bm{\hat{T}}_{t}\left( x,y \right) \right|}{\bm{T}_{t}\left( x,y \right)}}},
\end{equation}
d) the average of R square ($R^2$)
\begin{equation}\label{R2}
\setlength{\abovedisplayskip}{3pt}
\setlength{\belowdisplayskip}{3pt}
\begin{aligned}
& \overline{{{R}^{2}}}=\frac{1}{3\text{e}3}\sum\limits_{t=1}^{3000}{\left\{ 1-{\frac{\sum\limits_{\left( x,y \right)\in \Omega }{{\left[ \bm{T}_{t}\left( x,y \right)-\bm{\hat{T}}_{t}\left( x,y \right) \right]}^{2}}} {\sum\limits_{\left( x,y \right)\in \Omega }{{\left[ \bm{T}_{t}\left( x,y \right)-\bm{\bar{T}}_{t}\left( x,y \right) \right]}^{2}}}} \right\}}, \\
& \bm{\bar{T}}_{t}=\frac{1}{3\text{e}3}\sum\limits_{t=1}^{3000}{\bm{\hat{T}}_{t}}.
\end{aligned}
\end{equation}
For the first three  evaluation criteria $ \overline{RMSE} $, $ \overline{MAE} $ and $ \overline{MRE} $, the closer their values are to 0, the higher the accuracy of the trained model $\mathcal{M}\left(\bm{T}_{MP}, \bm{\tau}; \bm{\theta}\right)$. ${{R}^{2}}\to 1$ means that the trained model $\mathcal{M}\left(\bm{T}_{MP}, \bm{\tau}; \bm{\theta}\right)$ is accurate.

\subsection{Model accuracy analysis}
The prediction accuracy of five models $\mathcal{MQNN}_{{\varepsilon}_1}$, $\mathcal{MQNN}_{{\varepsilon}_2}$, $\mathcal{MQNN}_{{\varepsilon}_3}$, $\mathcal{MQNN}_{{\varepsilon}_4}$, and $\mathcal{MQNN}_{{\varepsilon}_5}$ are shown in TABLE \ref{model_acc_case1}. Apparently, three kinds of average errors, i.e., $ \overline{RMSE} $, $ \overline{MAE} $ and $ \overline{MRE} $, are sufficiently small for five models. The $\overline{R^2}$ of five models are greater than or equal to 0.99995. Therefore, the proposed Deep MC-QR method can reconstruct the temperature fields accurately.

\begin{table}[!htbp]
	\centering
	\caption{The prediction accuracy of five models trained by the MP temperature images with five kinds of noises.}
	\begin{tabular}{|c|c|c|c|c|c|}
		\hline
		Noise & Model & $ \overline{RMSE} $  & $ \overline{MAE} $ & $ \overline{MRE} $ & $\overline{R^2}$ \\
		\hline
		$ {\varepsilon}_1 $ & $\mathcal{MQNN}_{{\varepsilon}_1}$ & 0.07370 & 0.04932 & 0.00013 & 0.99996 \\
		\hline
		$ {\varepsilon}_2 $ & $\mathcal{MQNN}_{{\varepsilon}_2}$ & 0.07959 & 0.05066 & 0.00014 & 0.99996 \\
		\hline
		$ {\varepsilon}_3 $ & $\mathcal{MQNN}_{{\varepsilon}_3}$ & 0.07094 & 0.04489 & 0.00012 & 0.99996 \\
		\hline
		$ {\varepsilon}_4 $ & $\mathcal{MQNN}_{{\varepsilon}_4}$ & 0.08173 & 0.05263 & 0.00014 & 0.99995 \\
		\hline
		$ {\varepsilon}_5 $ & $\mathcal{MQNN}_{{\varepsilon}_5}$ & 0.08921 & 0.06056 & 0.00014 & 0.99995 \\
		\hline
	\end{tabular}
	\label{model_acc_case1}
\end{table}

\begin{figure}[!htbp]
	\setlength{\belowcaptionskip}{-0.3cm}
	\setlength{\belowcaptionskip}{-1cm} 
	\centering
	{\includegraphics[scale=0.56]{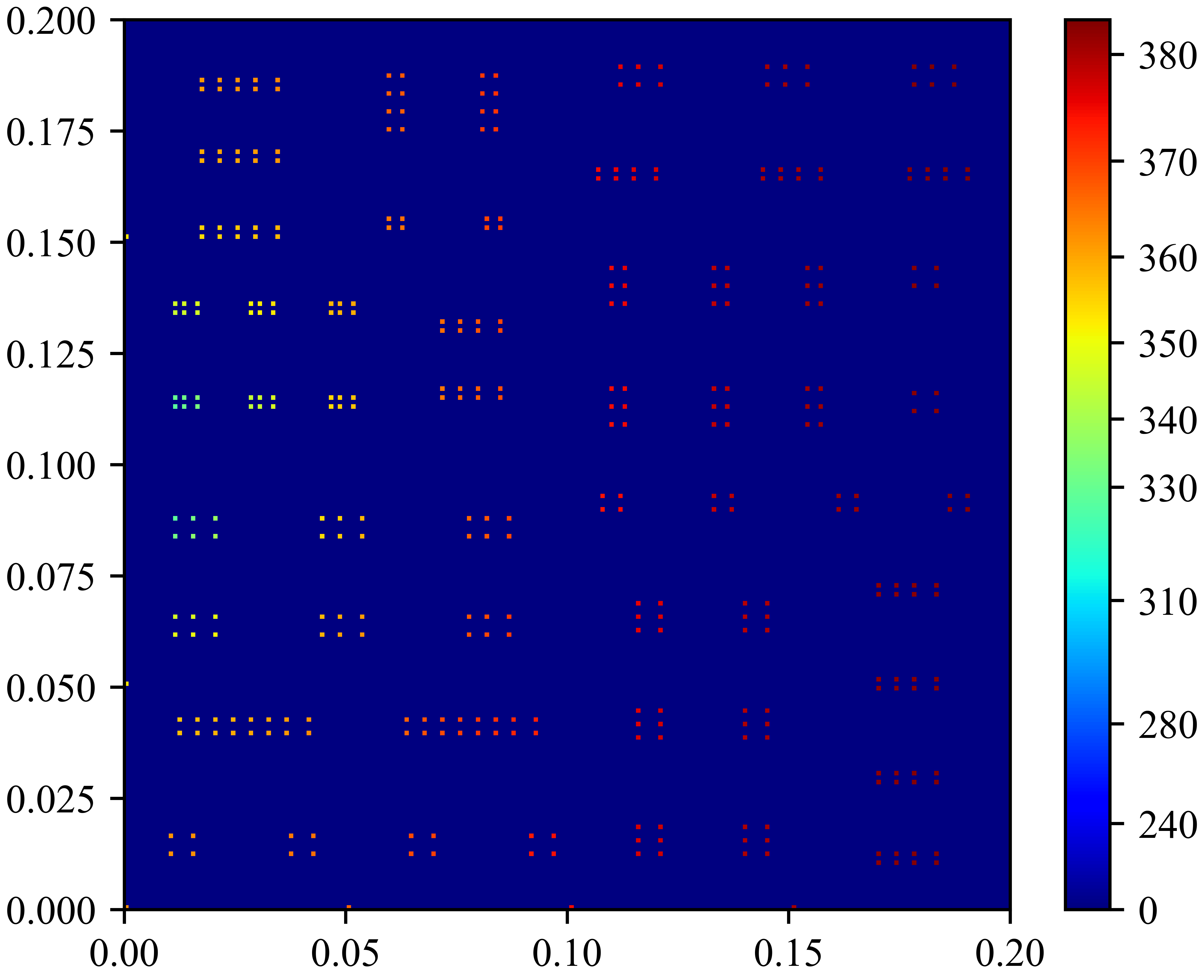}}
	\caption{A MP temperature image.}
	\label{MP1}
\end{figure}

\begin{figure}[!htbp]
	\setlength{\abovecaptionskip}{-0.03cm} 
	\centering
	{\includegraphics[scale=0.56]{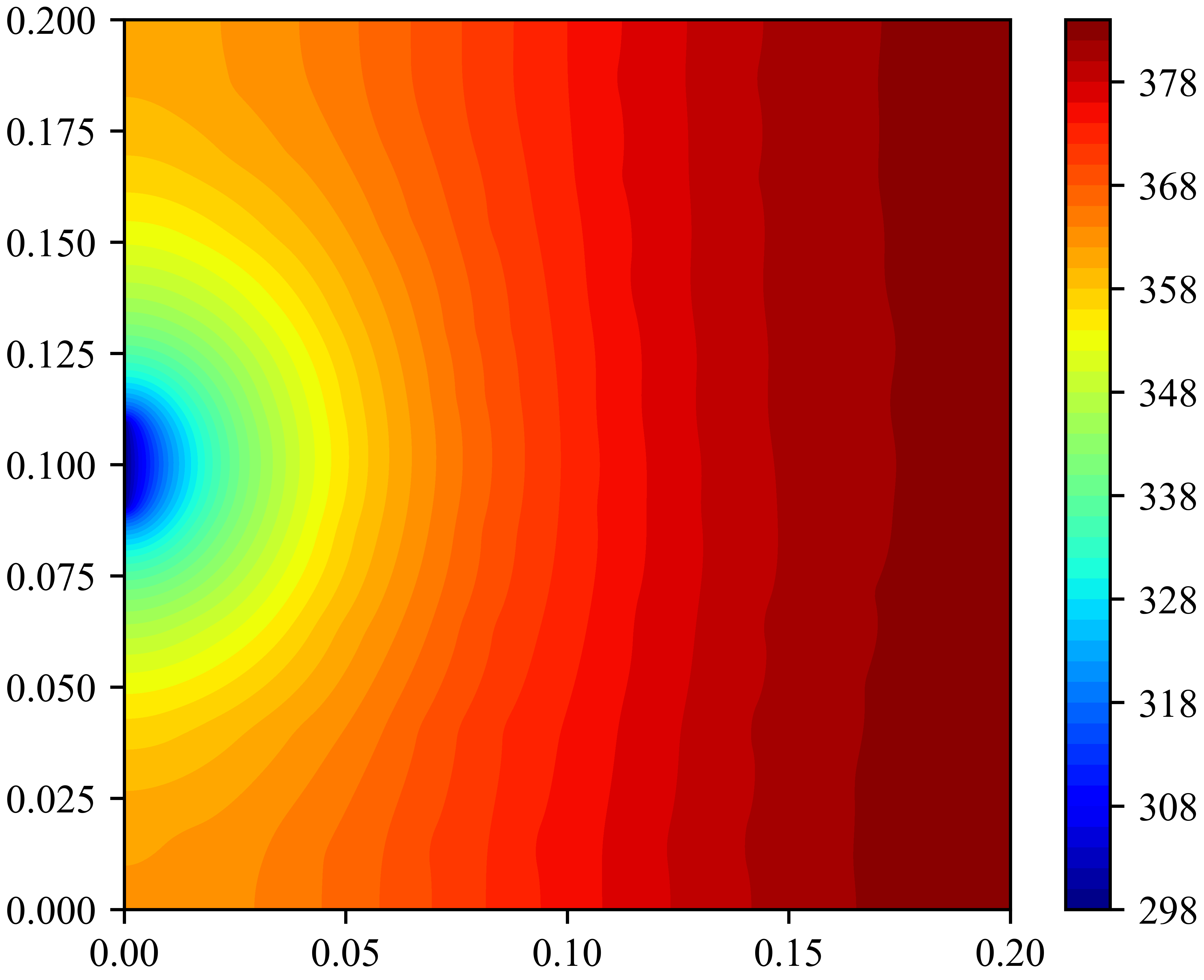}}
	\caption{Truth temperature field of the MP temperature image in Fig.\ref{MP1}.}
	\label{truth}
\end{figure}

\begin{figure}[!htbp]
	\setlength{\abovecaptionskip}{-0.03cm} 
	\centering
	{\includegraphics[scale=0.56]{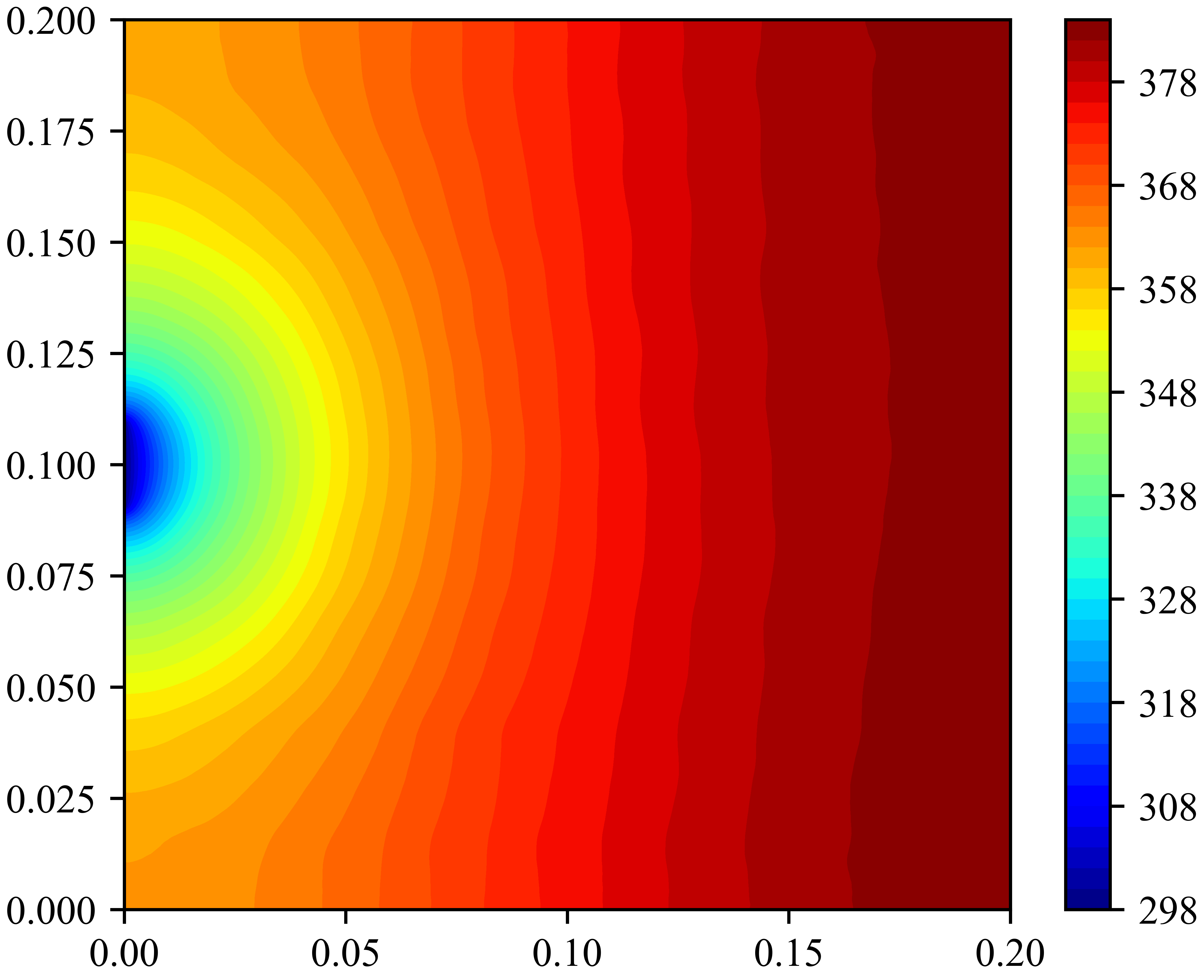}}
	\caption{Predicted temperature field by $\mathcal{MQNN}_{{\varepsilon}_1}$.}
	\label{prediction_1}
\end{figure}

\begin{figure}[!htbp]
	\setlength{\abovecaptionskip}{-0.03cm} 
	\centering
	{\includegraphics[scale=0.56]{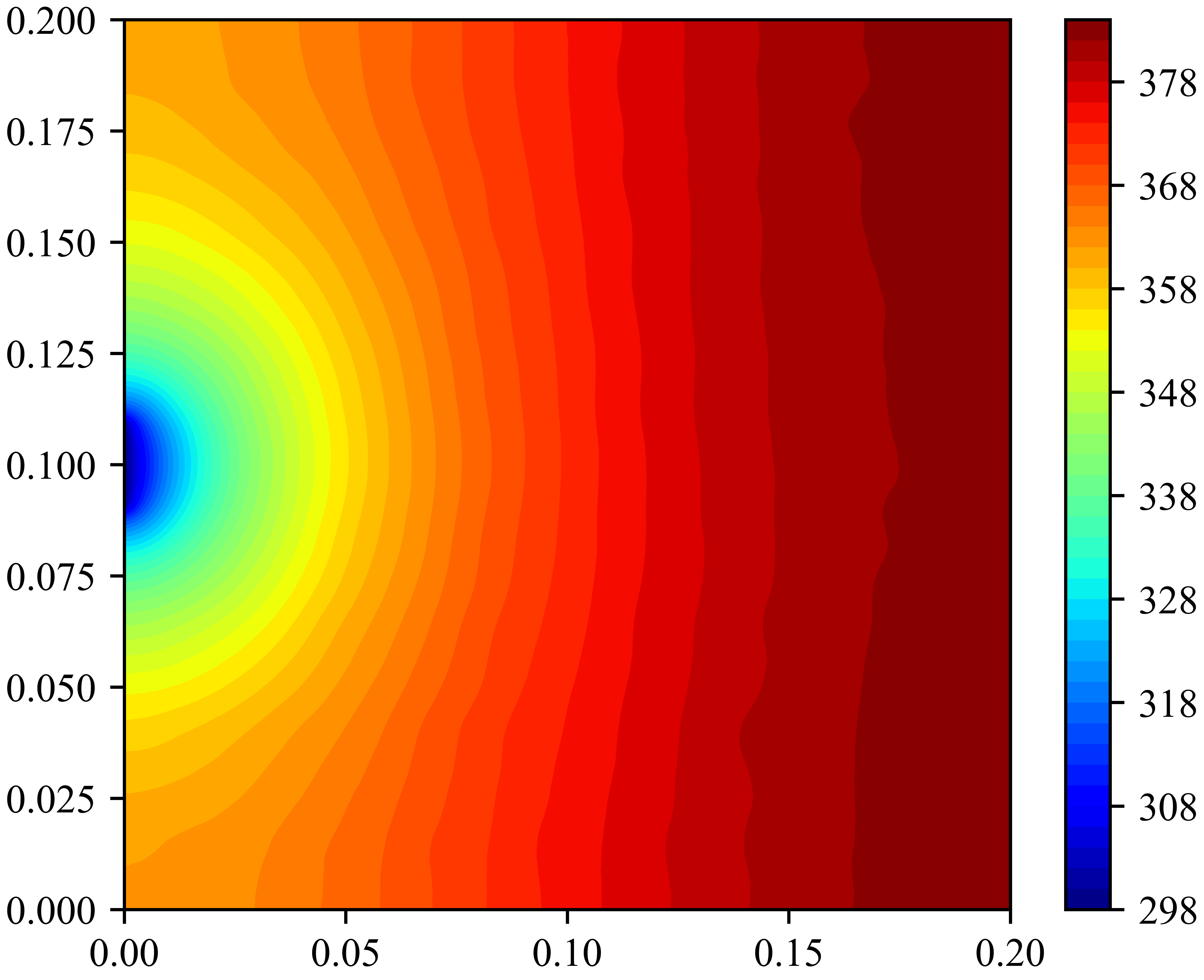}}
	\caption{Predicted temperature field by $\mathcal{MQNN}_{{\varepsilon}_2}$.}
	\label{prediction_2}
\end{figure}

\begin{figure}[!htbp]
	\setlength{\abovecaptionskip}{-0.03cm} 
	\centering
	{\includegraphics[scale=0.56]{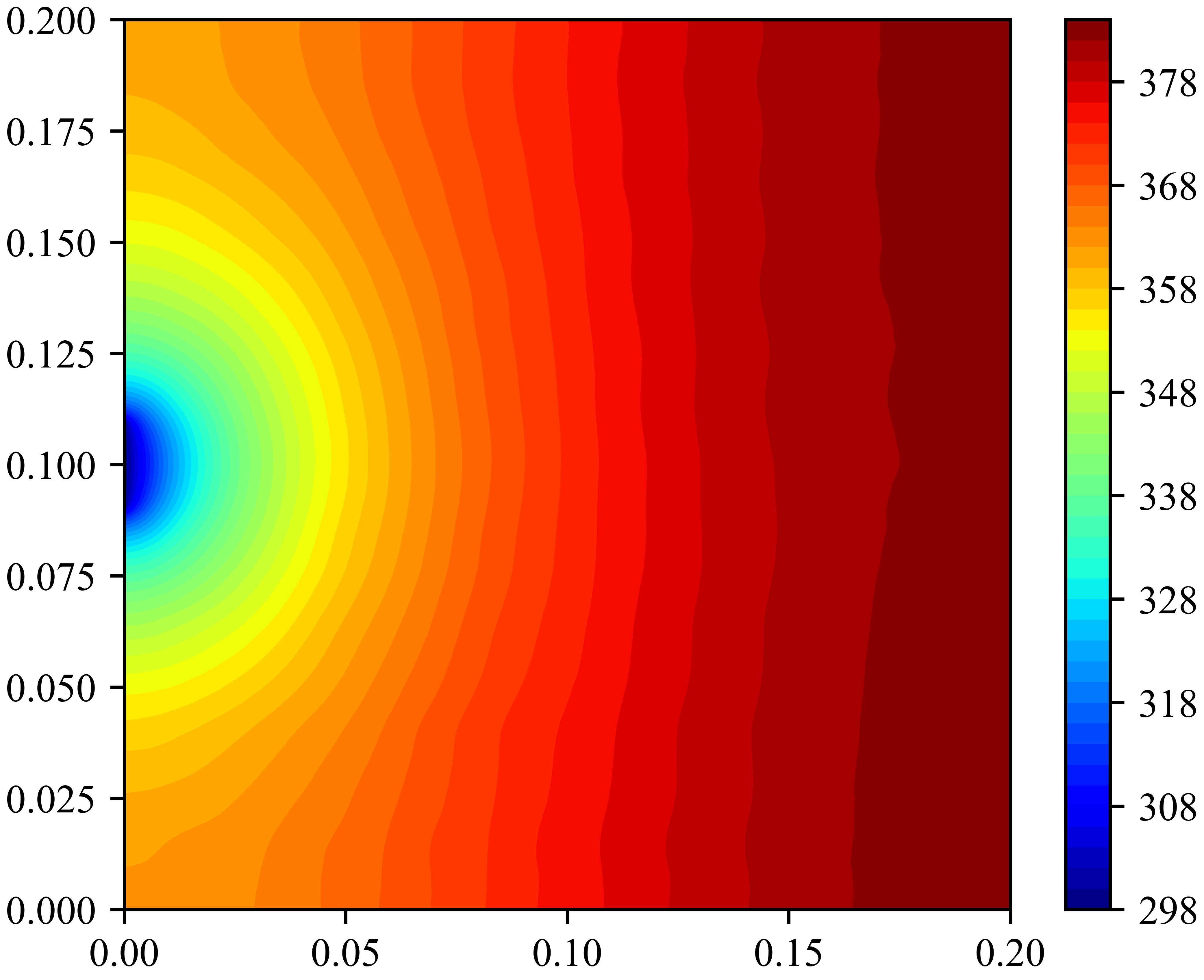}}
	\caption{Predicted temperature field by $\mathcal{MQNN}_{{\varepsilon}_3}$.}
	\label{prediction_3}
\end{figure}

\begin{figure}[!htbp]
	\setlength{\abovecaptionskip}{-0.03cm} 
	\centering
	{\includegraphics[scale=0.56]{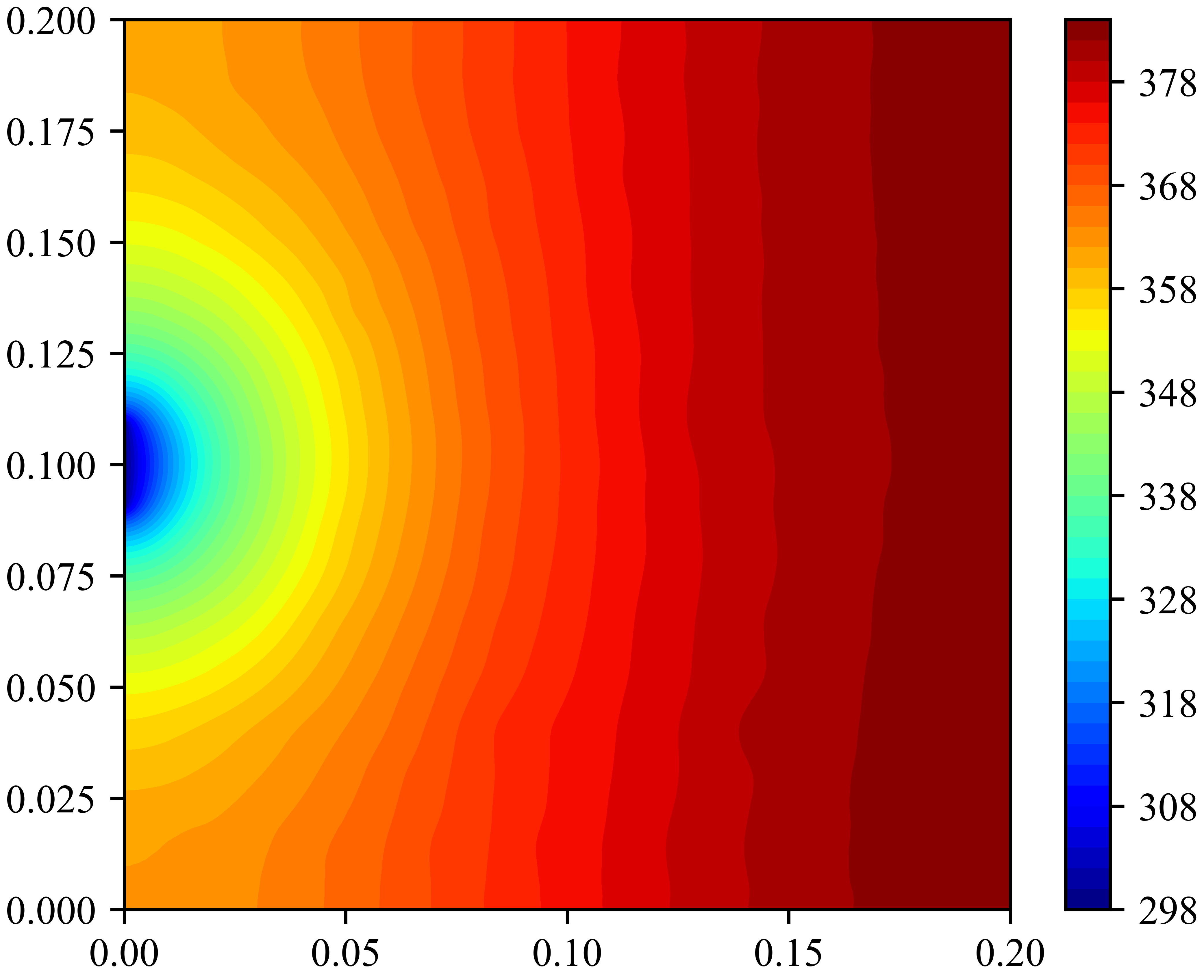}}
	\caption{Predicted temperature field by $\mathcal{MQNN}_{{\varepsilon}_4}$.}
	\label{prediction_4}
\end{figure}

\begin{figure}[!htbp]
	\setlength{\abovecaptionskip}{-0.03cm} 
	\centering
	{\includegraphics[scale=0.56]{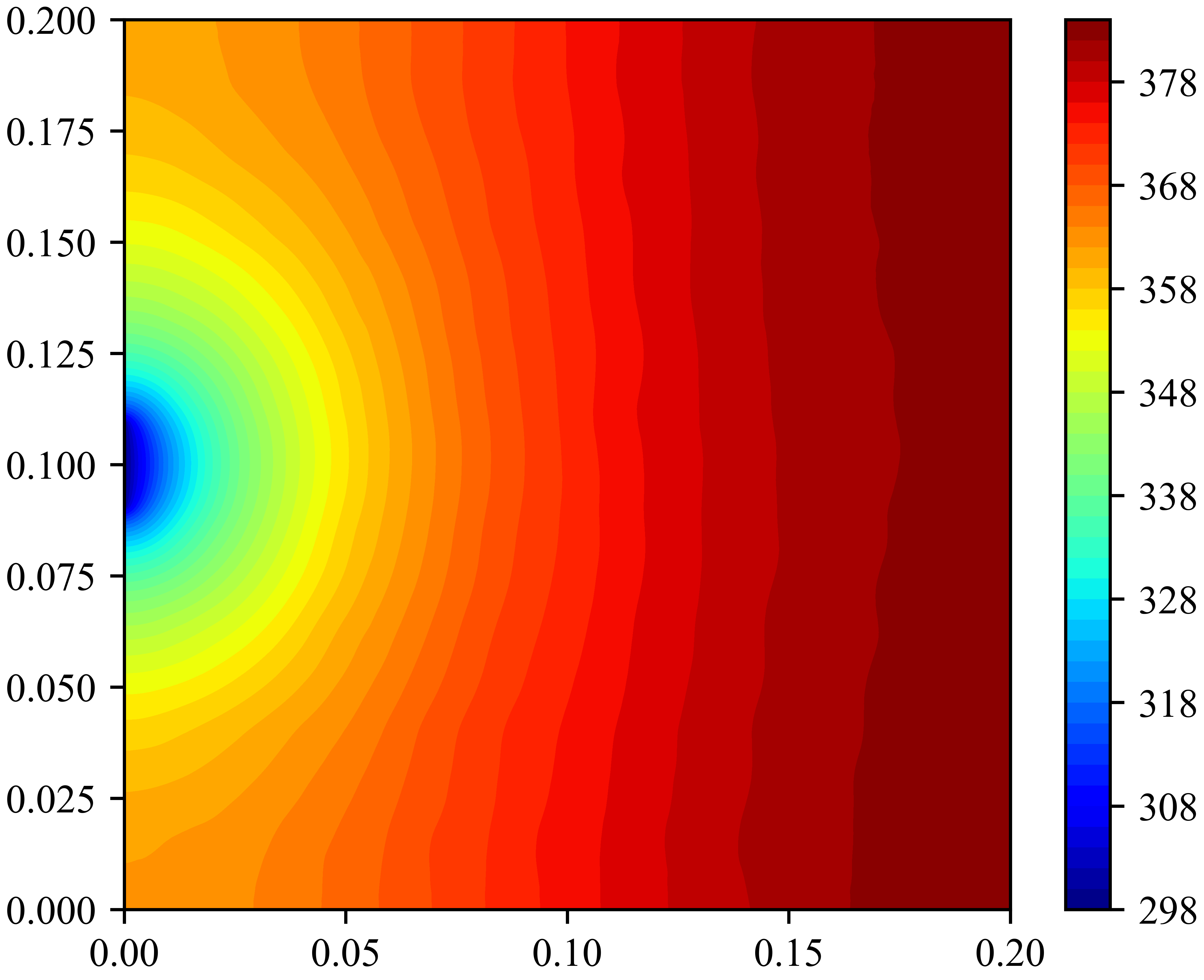}}
	\caption{Predicted temperature field by $\mathcal{MQNN}_{{\varepsilon}_5}$.}
	\label{prediction_5}
\end{figure}

This experiment randomly chooses a MP temperature image in Fig.\ref{MP1} (adding five kinds of data noises) to show the prediction performance of five models. Fig.\ref{truth} is the truth temperature field of the MP temperature image in Fig.\ref{MP1}. Based on five models $\mathcal{MQNN}_{{\varepsilon}_1}$, $\mathcal{MQNN}_{{\varepsilon}_2}$, $\mathcal{MQNN}_{{\varepsilon}_3}$, $\mathcal{MQNN}_{{\varepsilon}_4}$, and $\mathcal{MQNN}_{{\varepsilon}_5}$, five reconstructed temperature fields are presented in Fig.\ref{prediction_1}, Fig.\ref{prediction_2}, Fig.\ref{prediction_3}, Fig.\ref{prediction_4}, and Fig.\ref{prediction_5}, respectively. Compared with Fig.\ref{truth}, five reconstructed temperature fields are basically consistent with the truth temperature field.

\subsection{Aleatoric uncertainty quantification results analysis}
\begin{figure}[!htbp]
	\setlength{\belowcaptionskip}{-0.3cm}
	\centering
	{\includegraphics[scale=0.56]{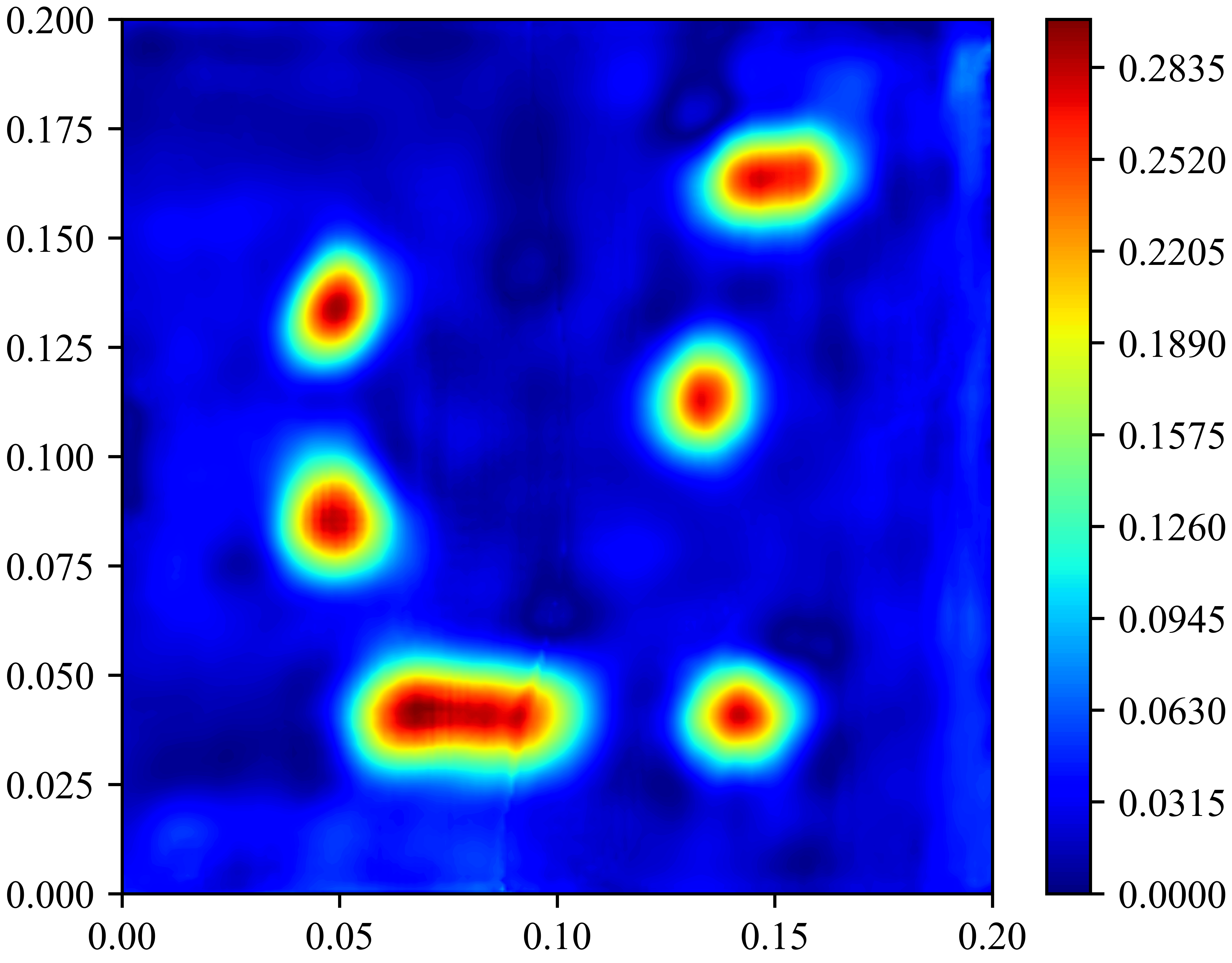}}
	\caption{Quantified aleatoric uncertainty by $\mathcal{MQNN}_{{\varepsilon}_1}$.}
	\label{alea1}
\end{figure}

\begin{figure}[!htbp]
	\setlength{\belowcaptionskip}{-0.3cm}
	\centering
	{\includegraphics[scale=0.56]{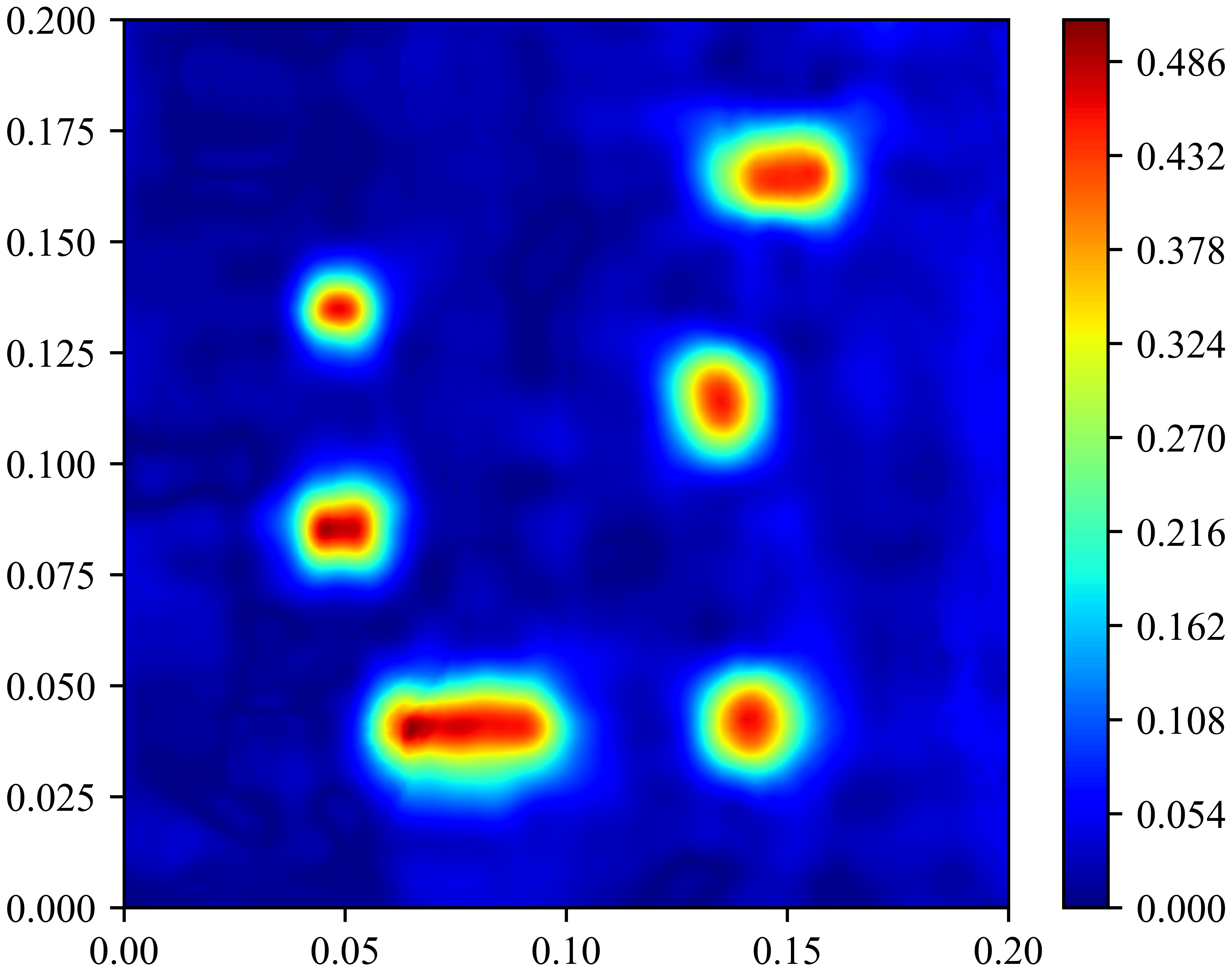}}
	\caption{Quantified aleatoric uncertainty by $\mathcal{MQNN}_{{\varepsilon}_2}$.}
	\label{alea2}
\end{figure}

\begin{figure}[!htbp]
	\setlength{\belowcaptionskip}{-0.3cm}
	\centering
	{\includegraphics[scale=0.56]{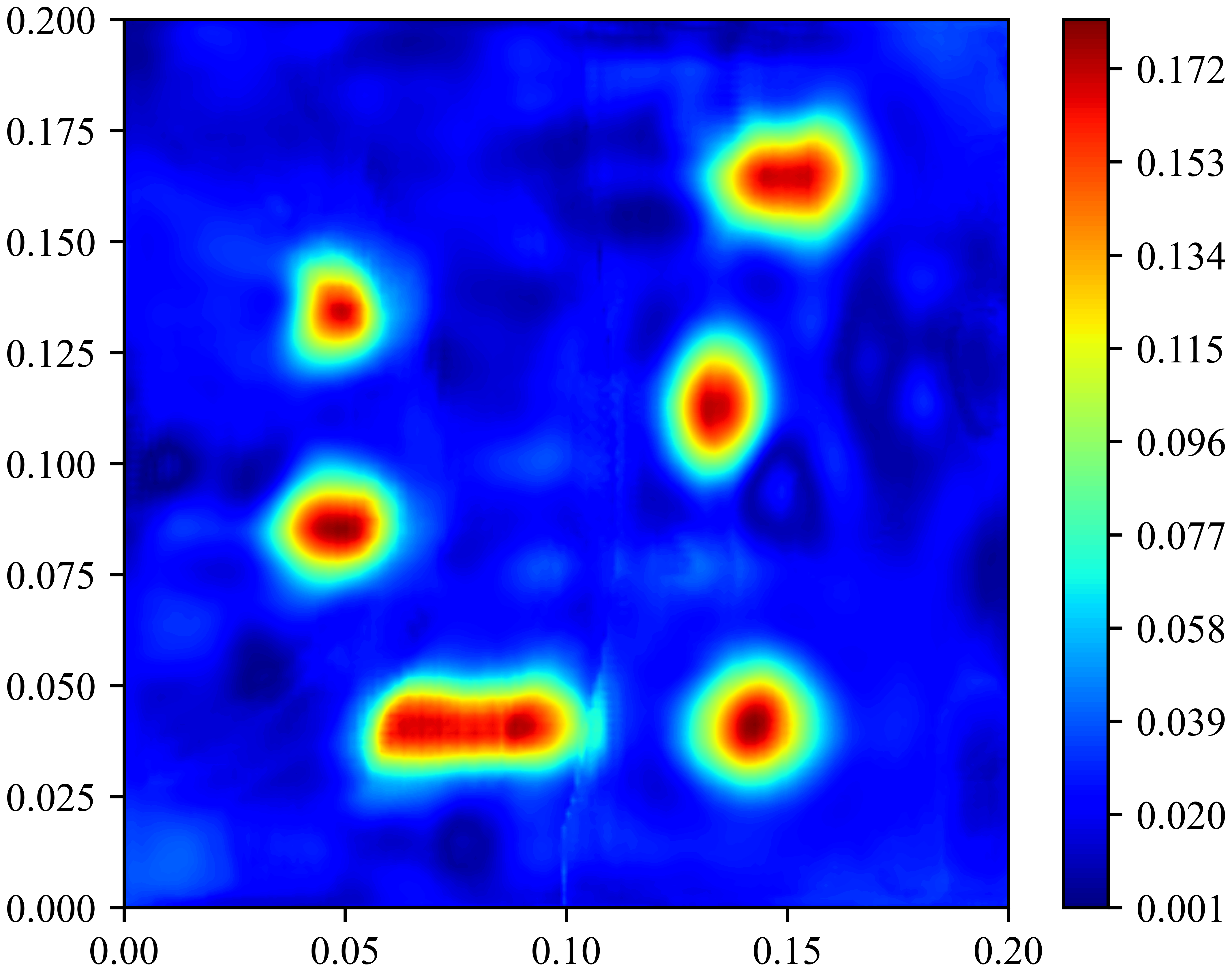}}
	\caption{Quantified aleatoric uncertainty by $\mathcal{MQNN}_{{\varepsilon}_3}$.}
	\label{alea3}
\end{figure}

\begin{figure}[!htbp]
	\setlength{\belowcaptionskip}{-0.3cm}
	\centering
	{\includegraphics[scale=0.56]{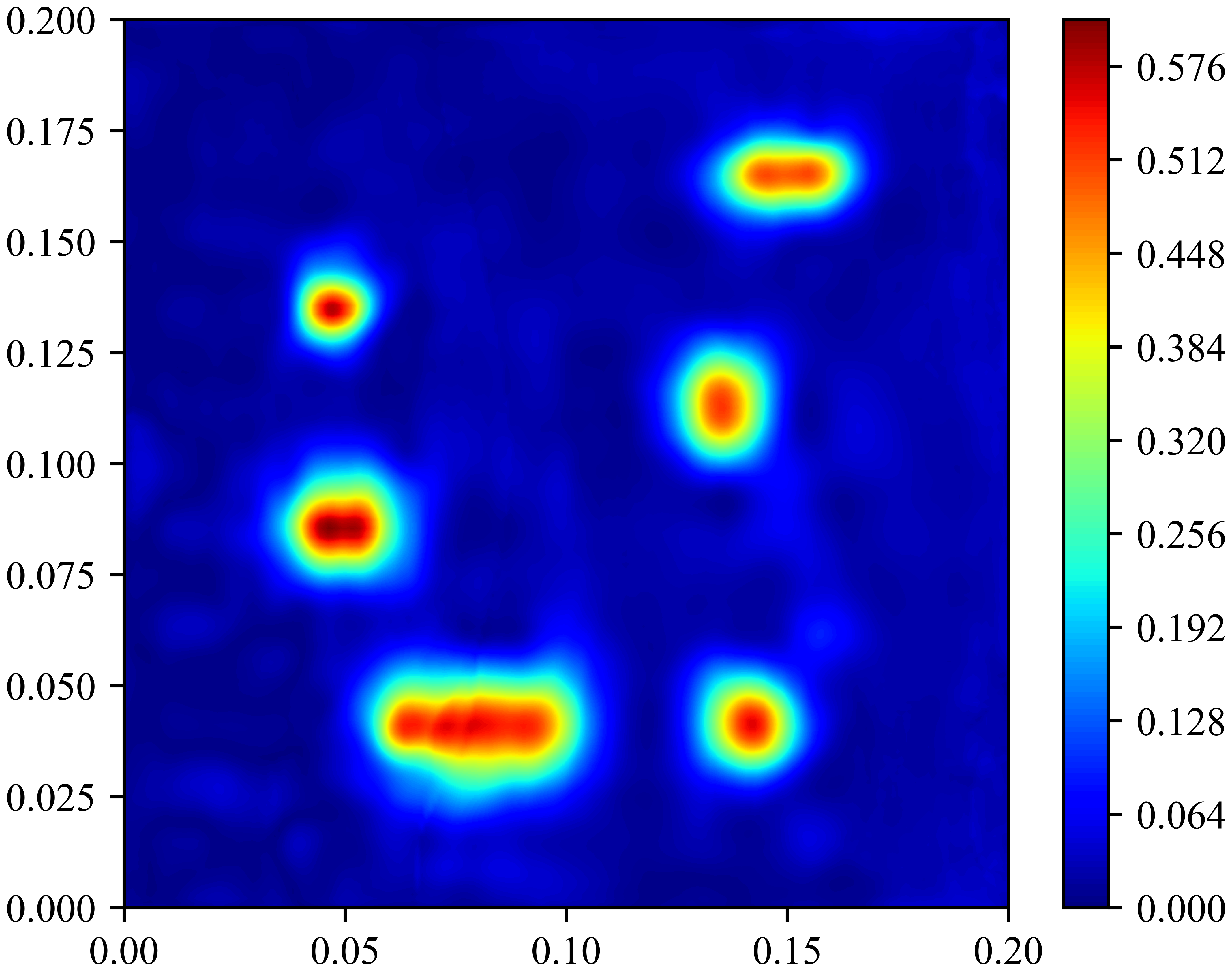}}
	\caption{Quantified aleatoric uncertainty by $\mathcal{MQNN}_{{\varepsilon}_4}$.}
	\label{alea4}
\end{figure}

For the MP temperature image in Fig.\ref{MP1}, the quantified aleatoric uncertainties by four models $\mathcal{MQNN}_{{\varepsilon}_1}$, $\mathcal{MQNN}_{{\varepsilon}_2}$, $\mathcal{MQNN}_{{\varepsilon}_3}$ and $\mathcal{MQNN}_{{\varepsilon}_4}$ are shown in Fig.\ref{alea1}, Fig.\ref{alea2}, Fig.\ref{alea3} and Fig.\ref{alea4}, respectively. According to section \ref{sec41}, these four models are trained by the datasets that include the Gaussian noises ($ {\varepsilon}_1 \sim N\left ( 0, {0.3}^2  \right )  $, $ {\varepsilon}_2 \sim N\left ( 0, {0.50}^2  \right )  $) and uniform noise ($ {\varepsilon}_3 \sim U\left ( -0.3, 0.3  \right )  $, $ {\varepsilon}_4 \sim U\left ( -1, 1  \right )  $) in the green points in Fig.\ref{MP_case}. Referring to Fig.\ref{alea1}, Fig.\ref{alea2}, Fig.\ref{alea3} and Fig.\ref{alea4}, the aleatoric uncertainties of the areas around the green points (Fig.\ref{MP_case}) are much larger than the other areas. Besides, the maximum values of two aleatoric uncertainties in Fig.\ref{alea1} and Fig.\ref{alea2} are approximately equal to 0.3 and 0.5, respectively. For the models $\mathcal{MQNN}_{{\varepsilon}_3}$ $\mathcal{MQNN}_{{\varepsilon}_4}$ trained by data with the uniform noises ($ {\varepsilon}_3$, $ {\varepsilon}_4  $), the quantified aleatoric uncertainties are approximately equal to the standard deviations 0.173 and 0.577 of two uniform noises, respectively. In summary, the proposed Deep MC-QR method can accurately quantify the aleatoric uncertainty caused by data noise.

\subsection{The effect of aleatoric uncertainty for TFR}
\begin{figure}[!htbp]
	\setlength{\abovecaptionskip}{-0.03cm}
	\centering
	{\includegraphics[scale=0.56]{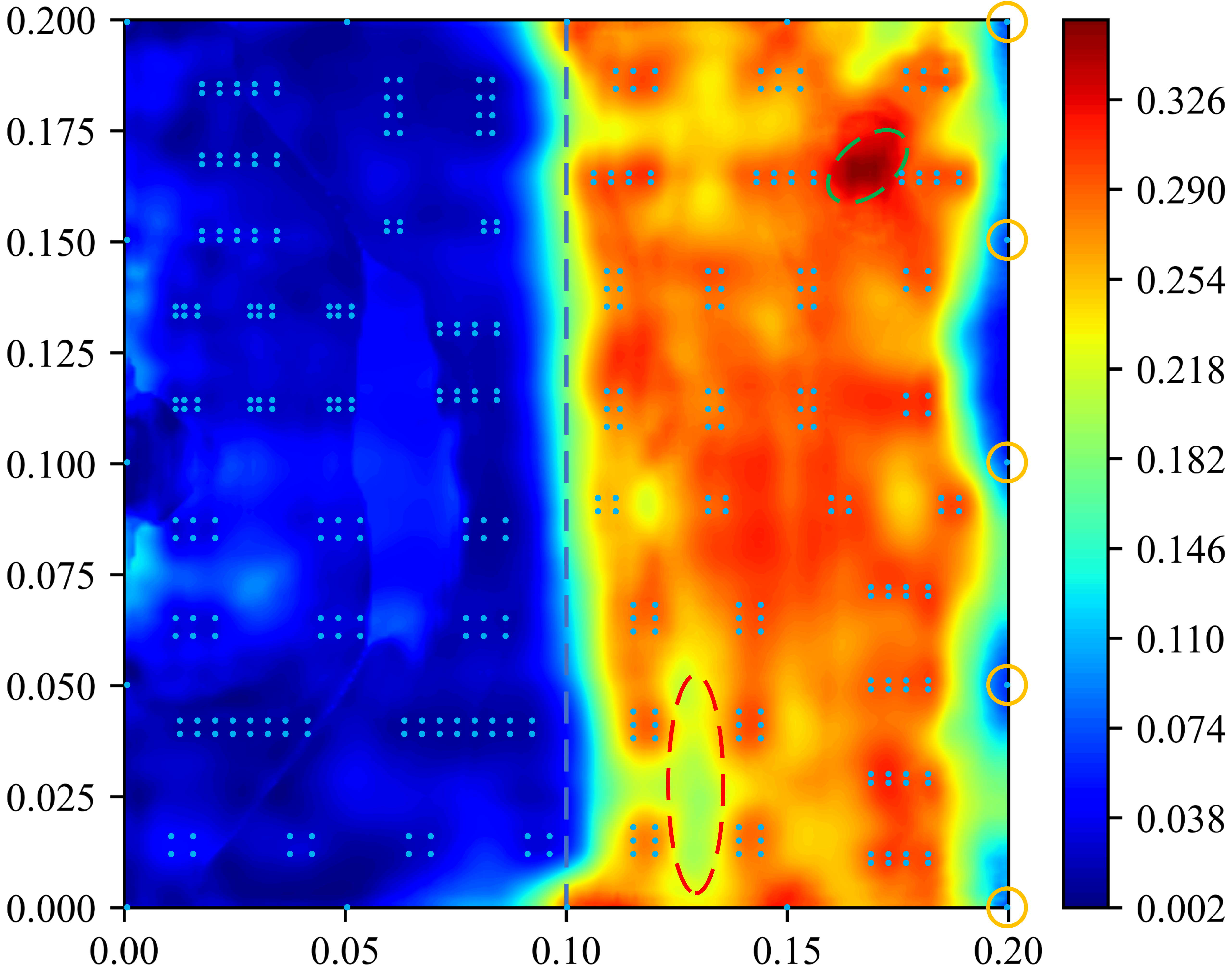}}
	\caption{The fusion image between the monitoring point positions (Fig.\ref{MP_case}) and the quantified aleatoric uncertainty by $\mathcal{MQNN}_{{\varepsilon}_5}$.}
	\label{noise_effect}
\end{figure}

For the model $\mathcal{MQNN}_{{\varepsilon}_5}$, it is trained by the data with the Gaussian noise $ {\varepsilon}_5 \sim N\left ( 0, {0.3}^2  \right )  $ for monitoring points to the right of the blue dotted line in Fig.\ref{MP_case}. The fusion image between the monitoring point positions (Fig.\ref{MP_case}) and the quantified aleatoric uncertainty by $\mathcal{MQNN}_{{\varepsilon}_5}$ is shown in Fig.\ref{noise_effect}.

Referring to Fig.\ref{noise_effect}, the aleatoric uncertainty corresponding to the area to the right of the blue dotted line is much larger than the area to the left of the blue dotted line. Besides, there are four especial phenomenons in Fig.\ref{noise_effect} as follows:

a) \textbf{The maximum aleatoric uncertainty (about 0.35) is larger than the actual data noise (0.3).} As shown in Fig.\ref{noise_effect}, the maximum value of quantified aleatoric uncertainty is approximately equal to 0.35, while the value of actual Gaussian noise is equal to 0.3.

b) \textbf{There is also aleatoric uncertainty on the non-monitoring point area to the right of the blue dotted line.} Compared with the non-monitoring point area to the left of the blue dotted line, the aleatoric uncertainty of the non-monitoring point area to the right of the blue dotted line is relatively large as shown in Fig.\ref{noise_effect}. However, the non-monitoring point area to the left of the blue dotted line does not contain data noise.

c) \textbf{The aleatoric uncertainty in the non-monitoring point area to the right of the blue dotted line has large differences.} As shown in Fig.\ref{noise_effect}, the aleatoric uncertainty in the green dashed ellipse is larger than in the red dashed ellipse for the non-monitoring point area. In fact, both non-monitoring point areas do not include data noise.

d) \textbf{The uncertainty around five monitoring points (in the orange circle) on the right boundary is small.}

In fact, the spread of data noise results in the above four especial phenomenons. Firstly, The data noise propagation leads to the accumulation of aleatoric uncertainty. Thereby, the maximum aleatoric uncertainty is larger than the actual Gaussian noise (phenomenon a)). Secondly, the aleatoric uncertainty is transferred to the surrounding area due to noise propagation, which brings about phenomenon b). Thirdly, the data noise may include positive or negative Gaussian noise. Thus, the accumulation of positive or negative Gaussian noise will lead to the phenomenon c). According to section \ref{sec41}, five monitoring points (in the orange circle) on the right boundary do not contain data noise. Therefore, the phenomenon d) appears in Fig.\ref{noise_effect}.

In summary, data noise causes the reconstructed temperature field to have aleatoric uncertainties at and around the monitoring point. Therefore, aleatoric uncertainty quantification is important to ensure the creditability of the reconstructed temperature field.

\section{Conclusions}
This paper proposes the Deep MC-QR method to reconstruct the temperature field and quantify the aleatoric uncertainty caused by data noise. By physical knowledge to guide the training of CNN, the proposed Deep MC-QR method can reconstruct an accurate TFR surrogate model without using any labeled training data. Besides, the proposed Deep MC-QR method can quantify aleatoric uncertainty by constructing a quantile level image for each MP temperature image. This paper adds five kinds of data noise in the MP temperature image. The results show that the proposed Deep MC-QR method can accurately reconstruct the temperature field and quantify the aleatoric uncertainty precisely. Especially, the results also show that data noise causes the reconstructed temperature field to have aleatoric uncertainties at and around the monitoring point. Thus, aleatoric uncertainty quantification is necessary for the TFR problem.

\section*{Acknowledgment}
This work was supported by the Postgraduate Scientific Research Innovation Project of Hunan Province (No.CX20200006) and the National Natural Science Foundation of China (Nos.11725211).

\bibliography{mybibfile}

\end{document}